\documentclass[runningheads]{llncs}
\usepackage{eccv}
\usepackage{eccvabbrv}
\usepackage{array}
\usepackage[accsupp]{axessibility}
\usepackage{graphicx}
\usepackage{booktabs}
\usepackage{multirow}
\usepackage{xspace}
\usepackage[dvipsnames]{xcolor}
\usepackage[breaklinks,colorlinks,citecolor=eccvblue,backref=page]{hyperref}
\usepackage{orcidlink}

\newcommand{\method}{DGE\xspace}
\newcolumntype{C}[1]{>{\centering\arraybackslash}p{#1}}

\DeclareMathOperator*{\argmin}{arg\,min}

\title{\method: Direct Gaussian 3D Editing by Consistent Multi-view Editing}
\titlerunning{\method: Direct Gaussian 3D Editing by Consistent Multi-view Editing}

\author{Minghao Chen \and
Iro Laina \and
Andrea Vedaldi}
\authorrunning{M.~Chen et al.}
\institute{Visual Geometry Group, University of Oxford \\
\email{\{minghao, iro, vedaldi\}@robots.ox.ac.uk} \\
\href{https://silent-chen.github.io/DGE/}{\tt\small {\nolinkurl{silent-chen.github.io/DGE}}}}

\begin{document}
\maketitle
\begin{abstract}

We consider the problem of editing 3D objects and scenes based on open-ended language instructions.
A common approach to this problem is to use a 2D image generator or editor to guide the 3D editing process, obviating the need for 3D data.
However, this process is often inefficient due to the need for iterative updates of costly 3D representations, such as neural radiance fields, either through individual view edits or score distillation sampling.
A major disadvantage of this approach is the slow convergence caused by aggregating inconsistent information across views, as the guidance from 2D models is not multi-view consistent.
We thus introduce the \emph{Direct Gaussian Editor} (\method), a method that addresses these issues in two stages.
First, we modify a given high-quality image editor like InstructPix2Pix to be multi-view consistent.
To do so, we propose a training-free approach that integrates cues from the 3D geometry of the underlying scene.
Second, given a multi-view consistent edited sequence of images, we \emph{directly} and efficiently optimize the 3D representation, which is based on 3D Gaussian Splatting.
Because it avoids incremental and iterative edits, \method is significantly more accurate and efficient than existing approaches and offers additional benefits, such as enabling selective editing of parts of the scene.
\end{abstract}

\section{Introduction}%
\label{sec:intro}

Recent breakthroughs in 2D and 3D generation~\cite{rombach22high-resolution, ho2020denoising, poole2022dreamfusion, melas-kyriazi23realfusion} have significantly advanced AI-based content creation and editing. This progress is particularly impactful for 3D content, which traditionally requires highly specialized skills and years of experience.
Consequently, these innovations are unlocking new creative possibilities for artists and non-professional users alike.

In this work, we consider the problem of editing 3D models based on textual instructions.
Recent progress in this area is largely due to the introduction of new radiance field representations for objects and scenes, such as NeRFs~\cite{mildenhall20nerf:}, which are both high-quality and flexible.
Because they are differentiable, radiance fields integrate well with applications of machine learning to image generation. 
A notable example is InstructPix2Pix (IP2P)~\cite{brooks2022instructpix2pix}, a denoising diffusion model~\cite{song21denoising,rombach22high-resolution} that can edit an image based on textual instructions~\cite{brooks2022instructpix2pix}.
A significant benefit of differentiable 3D representations, when combined with image generation models, is that they eliminate the need for 3D annotated data.
As a result, several works have repurposed such models to edit 3D content instead, primarily in two ways.
The first was pioneered by InstructNeRF2NeRF~\cite{haque23instruct-nerf2nerf:}, which utilizes IP2P to alternate between editing a rendered view of the 3D model and updating the latter by training on the edited images.
The second approach~\cite{zhuang23dreameditor:,kamata2023instruct,sella23vox-e:} is to update the model in a distillation-like fashion~\cite{poole2022dreamfusion}.
Both approaches rely on iterative mechanisms to incorporate the edits into the 3D model, so that a single edit often requires several minutes, or even hours, to complete.

In this paper, we introduce the \emph{Direct Gaussian Editor} (\method), a 3D editor that addresses the shortcomings of previous attempts.
We design our method with three goals in mind:
(i) high fidelity,
(ii) high efficiency, and
(iii) selective editing (\ie, editing only a specific part of the scene).
To achieve these goals, we propose to change both the 3D representation and the update mechanism.

As a representation, we employ 3D Gaussian Splatting (GS)~\cite{kerbl233d-gaussian}, a radiance field model that is notably orders of magnitude faster than NeRF and NeRF-adjacent models~\cite{chen22tensorf:,sun22direct,chan22efficient} for both rendering and gradient computation.
Additionally, GS offers another important advantage\@:
as an explicit 3D representation made of local 3D primitives, the Gaussians, it supports local edits easily and efficiently, as long as the relevant Gaussians can be identified.
In practice, one can easily identify the Gaussians by fusing the output of a 2D segmenter~\cite{kirillov23segment} from several rendered views of the scene.

While GS can significantly speed up the 3D editing process, it does not remove the bottleneck caused by the need to iterate several times between rendering, reconstruction, and evaluation of the underlying image-based diffusion model.  
The iterative methods used in prior works are slow because of the lack of view-consistent edits that could be used to update the 3D model coherently.
Since 2D editors like IP2P are \emph{monocular} and provide a distribution over all possible edits, the probability of drawing independently two or more consistent edits from different views is nearly zero.
As a result, iterative dataset updates or distillation are ways of incrementally reaching a ``multi-view'' consensus with respect to the application of the monocular editor to such views.

Our main contribution is a more efficient alternative to this slow iterative process.
We propose a method to sample \emph{multi-view consistent} edits such that the 3D model can be updated by \emph{directly} fitting it to the edited views.
Our approach is inspired by recent progress in video generation and editing and, in particular, methods that extend image generators to video without additional training~\cite{tokenflow2023, wu2023tune, khachatryan2023text2video,ceylan2023pix2video,qi2023fatezero}.
The key insight is that multiple views of the 3D model can be interpreted as an orbital video of a static scene generated by a moving camera, making multi-view consistency analogous to temporal consistency in video editing.
To achieve multi-view consistency we adopt the spatio-temporal attention mechanism used in the video editing literature and extend it with additional epipolar constraints, which we can enforce due to the 3D nature of our problem.
This method ensures that edits are coherent across multiple views, improving the efficiency and effectiveness of the 3D editing process.

Through qualitative and quantitative comparisons with prior works, we demonstrate two key advantages of \method, even compared to recent works such as the GaussianEditor~\cite{chen2023gaussianeditor}, which also uses Gaussian Splatting.
First, as a direct editing approach, \method %
results in a noticeable speed boost %
taking approximately 4 min for a single edit.
Second, ensuring multi-view consistent editing in the image space significantly simplifies the process of consolidating edits from various views into the 3D model.
This is reflected in both the number of updates required for convergence and the higher fidelity of the final result.

\section{Related Work}%
\label{sec:related_work}

\subsubsection{Image Editing.}

Due to the lack of large-scale training data to learn 3D editors directly, most such editors build on top of existing 2D image editors instead, so we discuss those first.
GLIDE~\cite{nichol21glide:} controls 2D image generation and editing using CLIP features~\cite{radford21learning}.
Methods like~\cite{gal2022textual,kumari2022customdiffusion} and DreamBooth~\cite{ruiz2023dreambooth} consider the problem of personalizing an image;
GLIGEN~\cite{li2023gligen} and others~\cite{chefer2023attendandexcite, chen2023trainingfree, epstein2023selfguidance} perform layout control, while ControlNet~\cite{Zhang_2023_ICCV} considers additional forms of control such as scribbles or depth maps.
DragDiffusion~\cite{shi23dragdiffusion:} and Drag-a-Video~\cite{teng23drag-a-video:} edit images and videos by dragging.
Others~\cite{meng2022sdedit, mokady2023null, bar2022text2live, Tumanyan_2023_CVPR, parmar2023zeroshot, kawar2023imagic, hertz2023prompttoprompt} cast image editing as image-to-image translation.
InstructPix2Pix (IP2P)~\cite{brooks2022instructpix2pix} fine-tunes Stable Diffusion~\cite{rombach22high-resolution} with image-and-text conditioning on a large synthetic dataset of language-driven edits.
The works of~\cite{Zhang2023MagicBrush, zhang2023hive} further improve InstructPix2Pix via manual labeling, while \cite{mirzaei23watch} aims at improving the localization of the edits.

\subsubsection{Ad-hoc 3D Editing.}

Several authors have explored various types of inputs and controls to edit 3D objects.
EditNeRF~\cite{liu21editing} updates shape and color in a radiance field based on user-provided scribbles,
3Designer~\cite{li223ddesigner:} and
SINE~\cite{bao23sine:} based on a single edited view,
SKED~\cite{mikaeili23sked:} based on 2D sketches,
Editable-NeRF~\cite{zheng23editablenerf:} based on  keypoints and
CoNeRF~\cite{kania2022conerf} based on attributes.
The work of~\cite{liu21editing} also considers sketch-based editing but for category-level 3D generators.
NeRF-Editing~\cite{yuan22nerf-editing:},
NeuMesh~\cite{yang22neumesh:} and
NeRFShop~\cite{jambon23nerfshop:} modify a radiance field based on meshes, NeuralEditor~\cite{chen23neuraleditor:} based on point clouds and \cite{xu22deforming} based on cages.
N3F~\cite{tschernezki22neural},
DFF~\cite{kobayashi2022decomposing},
SPIn-NeRF~\cite{mirzaei2023spin},
NeRF-in~\cite{liu22nerf-in:} and the work of~\cite{weder2023removing} consider segmenting and removing objects from a radiance field.
Control-NeRF~\cite{lazova23control-nerf:} allows both removing and moving objects.
Component-NeRF~\cite{lin23componerf:}
and the works of~\cite{zhang21editable,yang21learning} consider compositional editing.
Palette-NeRF~\cite{kuang23palettenerf:},
RecolorNeRF~\cite{gong2023recolornerf},
ICE-NeRF~\cite{lee2023ice} address recoloring and
ARF~\cite{zhang22arf:},
DeSRF~\cite{xu2023desrf},
StylizedNeRF~\cite{huang22stylizednerf:},
SNeRF~\cite{nguyen-phuoc22snerf:}
and the work of~\cite{chiang22stylizing,huang21learning} style transfer.
NeRFEditor~\cite{sun22nerfeditor:} integrates GAN-based stylization in NeRF\@.
Seal-3D~\cite{wang23seal-3d:} edits 3D scenes by learning interactive tools like brushes, deformation, and recoloring and Seal4D-NeRF~\cite{huang24seald-nerf:} extends it to dynamic NeRF models.
SceNeRFlow~\cite{tretschk23scenerflow:} allows for dynamic edits using DensePose for correspondence estimation~\cite{guler18densepose:}.

\subsubsection{Language-driven 3D Editing.}

Closer to our work, several authors have considered open-vocabulary or text-guided 3D editing.
Some authors use vision-language models based on CLIP~\cite{radford21learning,li2022languagedriven,li2022grounded} to edit or stylize a radiance field globally~\cite{wang22clip-nerf:,wang22nerf-art:,michel22text2mesh:,lei2022tango} 
or locally~\cite{song2023blending,kobayashi2022decomposing, wang2023inpaintnerf360, gordon2023blended, zhou2024feature}.
TextDeformer~\cite{gao23textdeformer:} focuses on manipulating only the shape of objects.
Some like
AvatarCLIP~\cite{hong22avatarclip:} consider animations as well.
Most recent works have shifted towards instruction-guided editing employing diffusion-based image generators like Stable Diffusion~\cite{rombach22high-resolution} or editors like Instruct-Pix2Pix~\cite{brooks2022instructpix2pix}.
There exist two main editing mechanisms among them. 
Instruct-NeRF2NeRF~\cite{haque23instruct-nerf2nerf:},
InstructP2P~\cite{xu23instructp2p:},
ProteusNeRF \cite{wang23proteusnerf:},
Edit-DiffNeRF~\cite{yu2023edit}, and
GaussianEditor~\cite{chen2023gaussianeditor, wang2024gaussianeditor}
repeatedly edit rendered views of the 3D object and update the 3D model accordingly, a process which is referred to as iterative dataset updates.
Instead, Instruct3Dto3D~\cite{kamata2023instruct},
DreamEditor~\cite{zhuang23dreameditor:},
Vox-E~\cite{sella23vox-e:},
ED-NeRF~\cite{park2023ed},
FocalDreamer~\cite{li2023focaldreamer},
Progressive3D~\cite{cheng2023progressive3d}
and~\cite{zhang2023text,zhou2023repaint}
update the 3D model using score distillation sampling (SDS).
While these works rely on costly, scene-specific optimization of radiance fields, SHAP-EDITOR~\cite{chen24shap-editor} learns a fast, feed-forward editor in the latent space of generative models but requires retraining for each new set of instructions.
Our editor is optimization-based, but significantly reduces the number of iterations required, up to a single one. 

Closely related to our work, ViCa-NeRF~\cite{dong24vica-nerf:} focuses on multi-view consistent editing. However, different from our method, which takes inspiration from video models, ViCA-NeRF leverages the 3D model's depth information to project features from key views into others using a blending module. 
As a result, it cannot handle edits that change the geometry of the original 3D shape.

Furthermore, methods like Vox-E~\cite{sella23vox-e:} and FocalDreamer~\cite{li2023focaldreamer} consider local 3D editing, but struggle to handle both local and global ones;
in contrast, our \method allows precise editing of small or large regions of the 3D scene.

\section{Preliminaries}%
\label{sec:Preliminaries}%

\subsubsection{Gaussian Radiance Fields.}%
\label{s:gs}

\newcommand{\x}{\boldsymbol{x}}
\newcommand{\y}{\boldsymbol{y}}
\newcommand{\z}{\boldsymbol{z}}
\newcommand{\w}{\boldsymbol{w}}
\newcommand{\bu}{\boldsymbol{u}}
\newcommand{\bmu}{\boldsymbol{\mu}}
\newcommand{\bnu}{\boldsymbol{\nu}}
\newcommand{\I}{\mathcal{I}}
\newcommand{\J}{\mathcal{J}}

A \emph{radiance field} is a pair of functions
$
\sigma : \mathbb{R}^3 \rightarrow \mathbb{R}^+
$
and
$
c : \mathbb{R}^3 \times \mathbb{S}^2 \rightarrow \mathbb{R}^3
$
mapping 3D points $\x \in \mathbb{R}^3$ to opacities $\sigma(\x)$ and directional colors $c(\x,\bnu)$, where $\bnu \in \mathbb{S}^2$ is a unit vector expressing the viewing direction.
An image $I$ of the radiance field is obtained via the \emph{emission-absorption} equation:
\begin{equation}\label{eq:ea}
I(\bu)
=
\int_{0}^\infty
c(\x_t,\bnu) \sigma(\x_t)
e^{-\int_0^t \sigma(\x_\tau) d\tau}
dt,
\end{equation}
where the 3D point $\x_t = \x_0 - t \bnu$ sweeps the ray that connects the camera center $\x_0$ to the pixel $\bu$ along direction $-\bnu$ (where the 3D point is expressed in the reference frame of the world, not the camera).

Prior works have explored several representations for these functions, including
MLPs~\cite{mildenhall20nerf:},
voxel grids~\cite{sun22direct}
and low-rank factorizations of the latter~\cite{chen22tensorf:,chan22efficient}.
\emph{Gaussian Splatting} (GS)~\cite{kerbl233d-gaussian} proposes a particularly efficient representation as a mixture of Gaussians
$
\mathcal{G} = \{(\sigma_i,\bmu_i,\Sigma_i,c_i)\}_{i=1}^G
$,
where $\sigma_i \geq 0$ is the opacity, $\bmu_i\in\mathbb{R}^3$ is the mean, $\Sigma_i\in\mathbb{R}^{3\times 3}$ is the covariance matrix, and $c_i : \mathbb{S} \rightarrow \mathbb{R}^3$ the directional color of each Gaussian.
The Gaussian functions are given by:
\begin{equation}
g_i(\x)
=
\exp\left(
  -\frac{1}{2}(\x-\bmu_i)^\top\Sigma_i^{-1}(\x-\bmu_i)
\right).
\end{equation}
The directional colors are given by functions
$
[c_i(\bnu)]_j = \sum_{l=0}^L \sum_{m=-l}^l c_{ijlm} Y_{lm}(\bnu),
$
where $Y_{lm}$ are spherical harmonics and $c_{ijlm}\in\mathbb{R}$ are corresponding coefficients.
The Gaussian mixture then defines the opacity and color functions as:
\begin{equation}\label{eq:gs-rf}
\sigma(\x) = \sum_{i=1}^G \sigma_i g_i(\x),
~~~
c(\x,\bnu)
= \frac{
  \sum_{i=1}^G c_i(\bnu) \sigma_i g_i(\x)
  }{
  \sum_{i=1}^G \sigma_i g_i(\x)
  }.
\end{equation}
Importantly, the integral in \cref{eq:ea} under the model in \cref{eq:gs-rf} can be approximated very efficiently and in a differentiable manner, as noted in~\cite{kerbl233d-gaussian}.

\subsubsection{Diffusion-based Generators and Editors.}%
\label{s:diffusion}

A \emph{diffusion-based image generator} is a model that allows one to sample image $I \in \mathbb{R}^{3\times H\times W}$ from a conditional distribution $p(I \mid y)$.
Here the condition $y$ is often a textual prompt, but it can also be an image or a combination of both.
Let
$
I_k = \sqrt{1- \beta_k^2} I + \beta_k \epsilon
$
be a noised version of the image $I$ one wishes to sample, where
$
\beta
$
is a sequence monotonically increasing from 0 to 1 and $\epsilon \sim \mathcal{N}$ is normal i.i.d.~noise.
The diffusion model consists of a denoising neural network $\Phi$ that, given the noisy image $I_k$, the noise level $k$ and the conditioning $y$, estimates the noise
$
\hat \epsilon = \Phi(I, y, k).
$
The image $I$ is sampled by starting from pure noise ($\beta_T=1$) followed by iterative denoising of the signal until $I=I_0$ is obtained~\cite{ho2020denoising,song21denoising,vargas23denoising}.

Several authors \cite{meng2022sdedit, hertz2023prompttoprompt, parmar2023zeroshot} have suggested using the denoising network of an off-the-shelf image generator such as Stable Diffusion (SD)~\cite{rombach22high-resolution,podell23sdxl:} to modify an existing image according to a prompt $y$ instead of generating it from scratch, thus editing it.
However, this process is somewhat suboptimal; instead, we make use of editors specifically trained to apply such modifications.
An example is IP2P~\cite{brooks2022instructpix2pix}, which also uses diffusion but directly implements conditional a distribution $p(I' \mid I, y)$ where $I'$ is the edited image, $I$ is the original image, and $y$ is a prompt describing how $I'$ should be obtained from $I$.
IP2P is in itself derived from the SD model by finetuning it to training triplets $(I',I,y)$ which are automatically generated using Prompt-to-Prompt~\cite{hertz2023prompttoprompt}, a slow but training-free editing method.

\section{Method}%
\label{sec:method}

\newcommand{\G}{\mathcal{G}}
\newcommand{\editor}{\mathcal{E}}
\newcommand{\editormv}{\mathcal{E_{\operatorname{mv}}}}

In this section, we describe our method for text-guided 3D scene editing\,---\,the Direct Gaussian Editor (\method).
Our main objective is 3D editing that is (i) faithful to the text instructions (high fidelity), (ii) fast and efficient, and (iii) partial to specific scene elements when desired.
Next, we introduce our approach and discuss how these criteria are met.

\subsection{Direct Gaussian Editor}%
\label{sec:direct}

\method is motivated by the following key observation:
given several images rendered from an existing 3D model, if one can obtain \emph{multi-view consistent edits} in image space, then the 3D model can be updated \emph{directly} and efficiently by fitting it to the edited images.
This approach is an alternative to slow iterative techniques like SDS~\cite{poole2022dreamfusion,sella23vox-e:,li2023focaldreamer} and iterative dataset updates (IDU)~\cite{haque23instruct-nerf2nerf:,chen2023gaussianeditor,fang23gaussianeditor:} that work around inconsistent image-level edits.
Thus, the main challenge for enabling direct 3D editing is to improve image-based editors to produce several \emph{consistent} edits across a number of images. %

In more detail, consider an initial reconstruction of a 3D scene as a mixture of Gaussians $\G$ (\cref{sec:Preliminaries}) and consider a camera trajectory
$
(\pi_t)_{t=1}^T
$
consisting of $T$ viewpoints.
We first render the corresponding views
$
\mathcal{I} = (I_t)_{t=1}^T
$
where
$
I_t = \operatorname{Rend}(\G, \pi_t)
$
is obtained with \cref{eq:ea}.
Let $y$ be the textual prompt describing the desired edits and let
$
\mathcal{I}' = (I'_t)_{t=1}^T
$
be the images obtained by editing $\mathcal{I}$ as according to $y$.
We can then update the 3D model $\G$ by fitting it to
$
\mathcal{I}'
$
as %
\begin{equation}\label{eq:optimization}
\G'
=
\argmin_\G
\sum_{t=1}^T \| I'_t - \operatorname{Rend}(\G, \pi_t) \|
\end{equation}
This rendering loss is often used to reconstruct a 3D scene from multiple views, and here it is used to update it.
In practice, GS makes this optimization efficient.

The key question is how to obtain the edited views $\mathcal{I}'$.
A standard approach is to apply a diffusion-based image editor $\editor$, such as IP2P~\cite{xu23instructp2p:} to \emph{individual} views, obtaining edited views $\mathcal{I}'$.
However, each of these views would then be drawn \emph{independently} from the distribution
$
I'_t \sim p(I'_t \mid I_t, y).
$
Hence, even though the initial views $I_t$ are consistent by construction, the edited views would not be as this method disregards their statistical dependency.
As a result, updating the 3D model $\G$ according to \cref{eq:optimization} would fuse inconsistent views, yielding a blurry outcome.
We thus seek to modify the image editor to account for the mutual dependency of the views and approximately draw samples from the joint distribution
$
p(I'_1,\dots,I'_T \mid I_1,\dots,I_T, y)
$.

\begin{figure}[t]
    \centering
    \includegraphics[width=\linewidth]{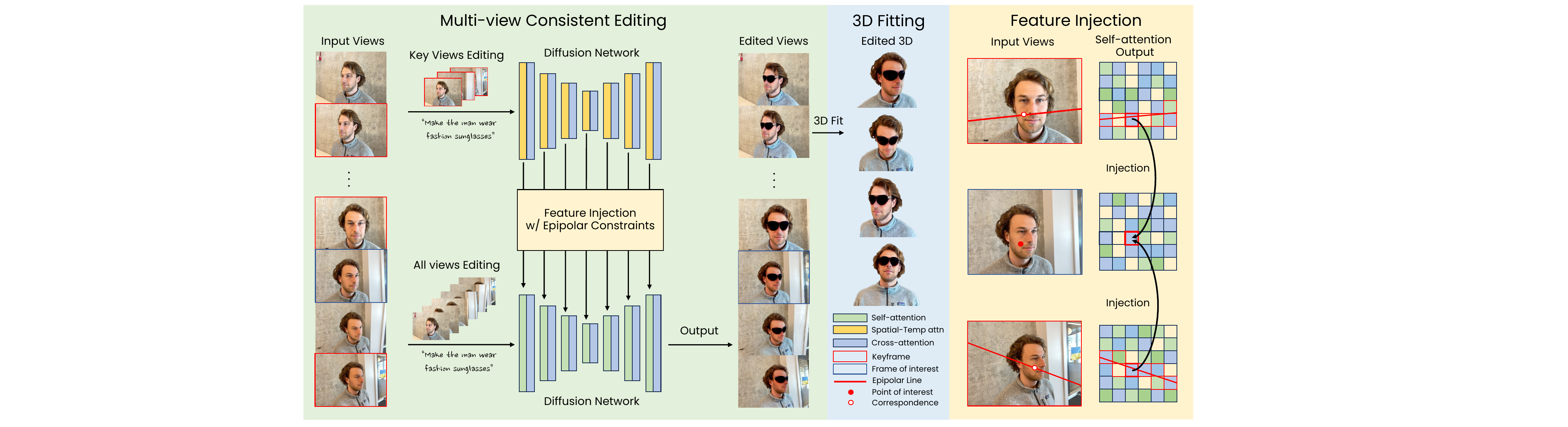}
    \caption{\textbf{Overview.} As shown on the left, our method is divided into two main parts: multi-view consistent editing with epipolar constraints and direct 3D fitting. In the multi-view editing stage, key views are randomly selected and jointly fed to the editing diffusion network to extract features with the \textit{spatial-temporal attention}. To edit other frames, the features of key views are injected into the diffusion network through correspondence matching on feature maps with epipolar constraints. The detailed feature injection process is shown on the right; only features with a red border (\ie, the points following epipolar constraints) are considered for correspondence matching.%
    }%
    \label{fig:overview}
\end{figure}

In summary, as shown in \Cref{fig:overview}, we divide the 3D editing process into two parts, detailed next: (a) multi-view consistent editing with epipolar constraints, and (b) the 3D reconstruction from edited images.

\subsection{Consistent Multi-view Editing with Epipolar Constraint}%
\label{sec:consistent_editing}

To achieve multi-view consistent edits, we render multiple views $\mathcal{I}$ from $\G$ and reorder them based on their camera positions to create a relatively smooth camera trajectory, forming a \emph{video}.
This approach leverages recent advances in video generation and editing, and in particular, the idea of extending image generators to video without additional training~\cite{tokenflow2023, wu2023tune, khachatryan2023text2video,ceylan2023pix2video,qi2023fatezero}. 
Treating the sequence $\mathcal{I}$ as a video of a static scene and applying these techniques to image \emph{editors}, consistency across views is analogous to temporal consistency in video editing.

Our multi-view editing process includes two sequential steps, key-view editing and feature injection. Specifically, during each denoising step, we first jointly edit select key views with \textit{spatio-temporal} attention.
Then, we inject the feature of the key views to other views using correspondences that are obtained from visual features guided by the epipolar constraint.

\subsubsection{Key-view editing.}
Let $\editor$ be a diffusion-based image editor that utilizes self-attention (\eg, SD-based).
Based on the above observations, 
we extend $\editor$ with a modified, \textit{spatio-temporal} version of self-attention, which jointly considers all images in $\mathcal{I}$, effectively turning the single-image editor into a multi-image editor $\editormv$.
Specifically, given queries $\{Q_t\}_{t=1}^T$, keys $\{K_t\}_{t=1}^T$, and values $\{V_t\}_{t=1}^T$ for each viewpoint $t$, the spatio-temporal attention block ensures that each frame attends to every other frame and is given by:
\begin{equation}
    \operatorname{STAttn}(Q, K, t) = \operatorname{Softmax}
    \left( \frac{ Q_t \cdot [\, K_1, \dots, K_T ]\ }{\sqrt{d}} \right) ,
\end{equation}
where $d$ is the embedding dimension of keys and queries.
The corresponding output features for frame $t$ are then computed as
$
\Phi_t = \operatorname{STAttn}(Q, K, t) \cdot [\, V_1, \dots, V_T ]\
$
(at each layer).
As the underlying image editor $\editor$, we utilize IP2P~\cite{brooks2022instructpix2pix}, which is a common choice in the literature \cite{haque23instruct-nerf2nerf:,chen2023gaussianeditor,wang2024gaussianeditor}.
However, the attention can be computationally expensive for longer sequences.
Therefore, $\editormv$ is only applied to select key views randomly selected from $\mathcal{T} = \{1, \dots, T\}$, where $T$ is the total number of images in the sequence.
In practice, we sample a random subset $\mathcal{K} \subset \mathcal{T}$ at each denoising iteration.
We detail the selection process in the supplement.
This step results in roughly consistent edits across all key views.

\subsubsection{Feature injection.} 
The goal of this step is to propagate the edited \emph{key view} features to \emph{all other} views to ensure the consistency of the edits.
Specifically, we inject the features by finding inter-frame correspondences with epipolar constraints. 
To find the correspondences between key views and other views, we first extract visual features $\Psi_t$ for all views in different layers of the denoising network $\Phi$, using the intermediate outputs of $\Phi$ (\ie, the inputs to each self-attention block). Then, point correspondences across views can be established by simply comparing their respective features at all spatial locations, and the key view features can be injected (based on these correspondences) into all remaining views, avoiding expensive self-attention blocks.

In addition, we can also leverage the fact that we can directly use 3D information to constrain the correspondence problem.
In fact, we have multiple views of the scene available, and the pose and calibration matrix of cameras are known.
This is without loss of generality since these assumptions are required to apply GS for scene reconstruction in the first place.
Therefore, we can estimate the fundamental matrix $F$ between two views and constrain the correspondence problem to points along an epipolar line.

Formally, given a feature map $\Psi_{t'}$ corresponding to image $I_{t'}$, where $t' \notin \mathcal{K}$, and the features $\Psi_{k^*}$  of a keyframe ${I_{k^*}}$, the correspondence map $M_{t'}$ is given by:
\begin{align}
\label{eq:correspondence}
M_{t'}[u] &= \argmin_{v,\, v^{\top}Fu=0} D\left( \Psi_{t'}[u],\, \Psi_{k^*}[v] \right), \quad \forall \, t' \in \mathcal{T} \backslash \mathcal{K}
\end{align}
where $D$ is the cosine distance, $u$ and $v$ index the feature maps spatially, $k^*$ is the index of the \emph{key} view that is the closest to view $t'$ (in terms of camera viewpoint), and $F$ is the fundamental matrix corresponding to the two views $t'$ and $k^*$.
$Fu$ is the epipolar line in view $k^*$ along which the corresponding point of $v$ in view $k^*$ must lie~\cite{hartley00multiple}.
Intuitively, using the epipolar constraint significantly reduces the search space of correspondences from a plane to spatial locations along a single line.
Given this constraint, the correspondence is decided based on the cosine distance of the features.
In practice, we compute the correspondence maps between a view $t'$ and its two nearest neighbors. For a given position $u$, we linearly combine the features of correspondences in the nearest two key views to obtain its edited feature.

This approach is similar in spirit to TokenFlow~\cite{tokenflow2023} but contributes two noteworthy improvements that are tailored to the 3D editing task in particular.
First, using IP2P instead of the 2D editor~\cite{Tumanyan_2023_CVPR} used in TokenFlow avoids the time-consuming DDIM~\cite{song2021denoising} inversion process.
Second, the epipolar constraint in \cref{eq:correspondence} allows us to leverage the 3D information that is available from the original model $\G$ when propagating edits.
This, in turn, can be particularly useful in cases where appearance is insufficient to compute correspondences (\eg, mostly uniform appearance, as shown in \cref{fig:epipolar}).

\subsection{Direct Reconstruction and Iterative Refinement}%
\label{sec:reconstruction}

After obtaining a set of edited images $\mathcal{I}$ with roughly consistent content, we can directly fit the Gaussian mixture $\G$ to the resulting images using \cref{eq:optimization}, without the need for distillation (SDS).
One of the notable advantages of 3D GS over NeRF is that it enables the rendering of images at full resolution and the application of robust image-level losses.
Specifically, we optimize $\G$ using LPIPS~\cite{zhang2018unreasonable} as an objective function between the edited $I'_t$ and the rendered $\hat{I}_t$ views in \cref{eq:optimization}.
The LPIPS loss is insensitive to small inconsistencies in the edited images, and helps to obtain a more consistent 3D output.

Finally, we observe that even though $\editormv$ results in generally consistent edits, some inconsistencies and artifacts may remain, especially around fine and detailed textures, which in turn lead to blurry results after updating $\G$.
We conjecture that this is due to the limited resolution of the feature space.
To address this issue, after obtaining $\G'$, we can again render images to be edited, repeating the process described in the previous section and re-updating the 3D model, similar to SDEdit~\cite{meng2022sdedit} or IM-3D~\cite{melas-kyriazi24im-3d}.

\subsection{Partial Editing}%
\label{ss:partial}

Because 3D GS is an \emph{explicit} 3D representation, if desired, a user can selectively edit a 3D scene by allowing only certain Gaussians to change, thus focusing on a specific region of interest while ensuring that the rest of the scene remains unchanged.
This is a significant advantage compared to implicit radiance field representations, where there is no direct correspondence between specific parameters and regions of space, so any update to the model tends to be global.

We thus propose to first obtain a mask for the Gaussians that should undergo editing. This could be done following existing works \cite{chen2023gaussianeditor, cen2023segment}. In this paper, we mainly adopt the approach in \cite{chen2023gaussianeditor}, which segments the 2D views first and unprojects the 2D segmentation results to 3D to obtain the masks of Gaussians.

We then follow the same steps outlined in \cref{sec:consistent_editing,sec:reconstruction}, but only optimize the reconstruction objective on regions defined by the rendered segmentation masks.
This training procedure leads to the partial change of a 3D model when a specific prompt is provided.

\section{Experiments}%
\label{sec:exp}

In this section, we first provide the implementation details of our method followed by qualitative and quantitative comparison with other methods.
We then provide an ablation study of our method.

\paragraph{Implementation Details.}

As our image editor, we use InstructPix2Pix~\cite{brooks2022instructpix2pix} for a fair comparison to prior methods using the same model. %
We use scenes from IN2N~\cite{haque23instruct-nerf2nerf:} and other real-world datasets, including LLFF and Mip-NeRF360~\cite{mildenhall2019llff, barron22mip-nerf} to demonstrate the ability of our method to edit 3D models.
We use 3D GS~\cite{kerbl233d-gaussian} as the 3D representation.
For editing, we use 20{-}30 views, which are edited according to the approach outlined in \cref{sec:consistent_editing} and then train $\G'$ with 500{-}1500 iterations depending on the complexity of the scenes.
We use LPIPS and L1 loss to train the 3D GS as suggested in IN2N~\cite{haque23instruct-nerf2nerf:}.
We also use classifier-free guidance to control the effect strength of the editing.
Most of the edits use the default setting, 7.5 for textual conditions and 1.5 for the image condition.
We perform the iterative refinement every 500 iterations.
We provide more details in the supplementary material.

\begin{table} [t!]
  \footnotesize
  \renewcommand*{\arraystretch}{0.95}
  \newcommand{\xpm}[1]{{\tiny$\pm#1$}}
  \centering
\setlength{\tabcolsep}{2.7pt}
\begin{tabular}{@{}lcccc@{}}
  \toprule
  \multirow{2}{*}{Method} & 3D & CLIP & CLIP Directional  & Avg. Editing \\
  & Model & Similarity & Similarity  &  Time \\
    \midrule
    Instruct-N2N~\cite{haque23instruct-nerf2nerf:} & NeRF & 0.215 & 0.064 & $\sim$ 51min \\
    ViCA-NeRF~\cite{dong24vica-nerf:} & NeRF & 0.204 & 0.044 & $\sim$ 28min \\
    \midrule
    GaussianEditor~\cite{chen2023gaussianeditor} & GS & 0.201 & 0.060 & $\sim$ 7min \\
    IP2P~\cite{xu23instructp2p:} + SDS~\cite{poole2022dreamfusion} & GS & 0.206 & 0.061 & $\sim$ 6min \\
    \midrule
    Ours & GS & \textbf{0.226} & \textbf{0.067} & $\sim$ 4min \\
  \bottomrule
\end{tabular}
\vspace{0.5em}
\caption{\textbf{Comparison with other editing methods.} Methods based on Gaussian Splatting are much faster than NeRF-based ones. Our \method achieves the best performance at almost half the time compared to GaussianEditor\@.}%
\label{table:comparison}
\end{table}

\paragraph{Evaluation.}

We provide both qualitative and quantitative evaluations of our method.
For quantitative evaluations, we follow common practice~\cite{haque23instruct-nerf2nerf:,chen2023gaussianeditor}.
We evaluate the alignment of edited 3D models and target text prompts with CLIP similarity score, \ie, the cosine similarity between the text and image embeddings encoded by CLIP\@.
Specifically, we randomly sample 20 camera poses from the training dataset of the 3D models and measure the CLIP similarity between the rendered images and the target text prompt.
Additionally, to measure the editing effect, we use the CLIP directional similarity, \ie, the cosine similarity between the image and text editing directions (target embeddings minus
source embeddings).
We evaluate all methods on 3 different scenes and 10 different prompts.
The detailed scene-prompt pairs are provided in the supplement.

\subsection{Comparisons with Prior Work}

We first compare our method with representative text-guided 3D editing methods, IN2N~\cite{haque23instruct-nerf2nerf:} and GaussianEditor~\cite{chen2023gaussianeditor}.
They both employ the iterative dataset updates (IDU) approach, while the former builds on a NeRF representation and the latter on Gaussian Splatting.
To edit a 3D model, they both iterate between the 3D model training and image editing.
Every 10 steps, they use the same 2D editor (IP2P) to edit the rendered image, replace the edited image in the 3D model training dataset, and continue the model training.
In addition, we include a baseline using the same GS framework as in our approach, but applying edits via SDS with InstructPix2Pix (IP2P + SDS).
This baseline enables a direct and fair comparison between our consistent multi-view editing approach and the popular SDS-based approach. Lastly, we compare with ViCA-NeRF~\cite{dong24vica-nerf:}, which also focuses on editing different views consistently. Different from ViCA-NeRF, which uses 3D model depth information to project features from key views into others using a blending module, our approach incorporates the constraints from epipolar lines to avoid depth estimation errors.

In \Cref{table:comparison} we compare \method to the above methods and baselines in terms of CLIP similarity score and CLIP directional similarity score. 
Our approach outperforms all prior work while cutting the editing time to almost half compared to GaussianEditor, the second fastest alternative.

\begin{figure}[t]
    \centering
    \includegraphics[width=\linewidth]{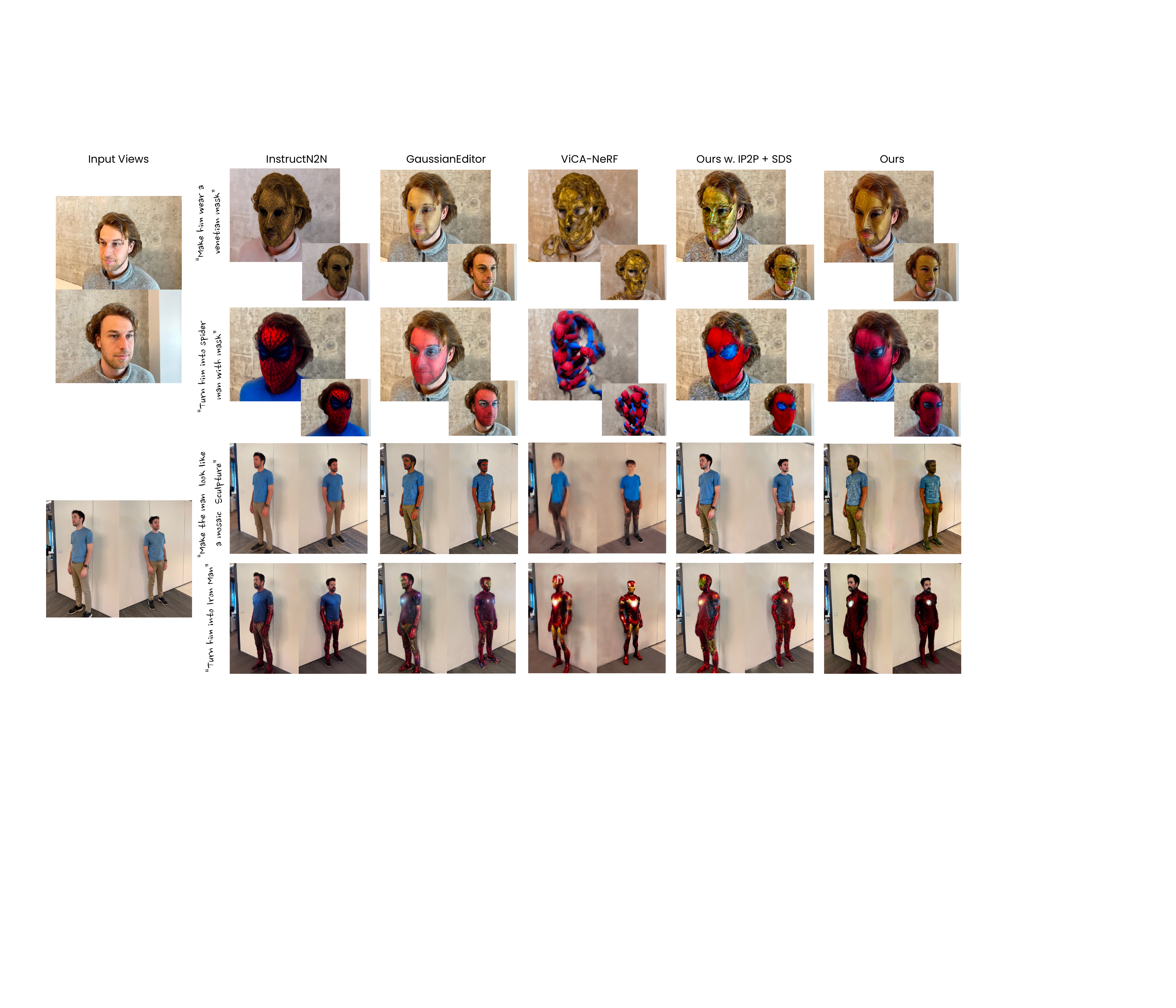}
    \caption{Comparison with other methods.
    Our method can provide fast and detailed editing effects, such as the textures on the Venetian mask and mosaic sculpture.
    Other methods, such as InstructN2N and IP2P+SD, fail to get the mosaic effects because they average over inconsistent editing.}
    \label{fig:main_comparison}
\end{figure}

\begin{table} [t!]
  \footnotesize
  \renewcommand*{\arraystretch}{0.95}
  \newcommand{\xpm}[1]{{\tiny$\pm#1$}}
  \centering
\setlength{\tabcolsep}{2.7pt}
\begin{tabular}{@{}cccc@{}}
  \toprule
  Multi-view  & Epipolar  & CLIP  & CLIP Directional  \\
  Editing & Constraints & Score & Score \\
    \midrule
    {} &  {} & 0.203 & 0.059 \\
    {} & \checkmark & 0.221 & 0.064 \\
    \checkmark & \checkmark & \textbf{0.226} & \textbf{0.067} \\
  \bottomrule
\end{tabular}
\vspace{0.5em}
\caption{\textbf{Ablation study on different components of our methods.} Without multi-view consistency editing, both the CLIP and CLIP directional scores drop significantly. Although the difference between with and without epipolar constraints is small, we provide a visual comparison in \cref{fig:epipolar} showing that it helps to improve the detailed texture especially when the appearance is relatively similar in different places.}%
\label{table:ablation}
\end{table}

\begin{figure}[t]
    \centering
    \includegraphics[width=\linewidth]{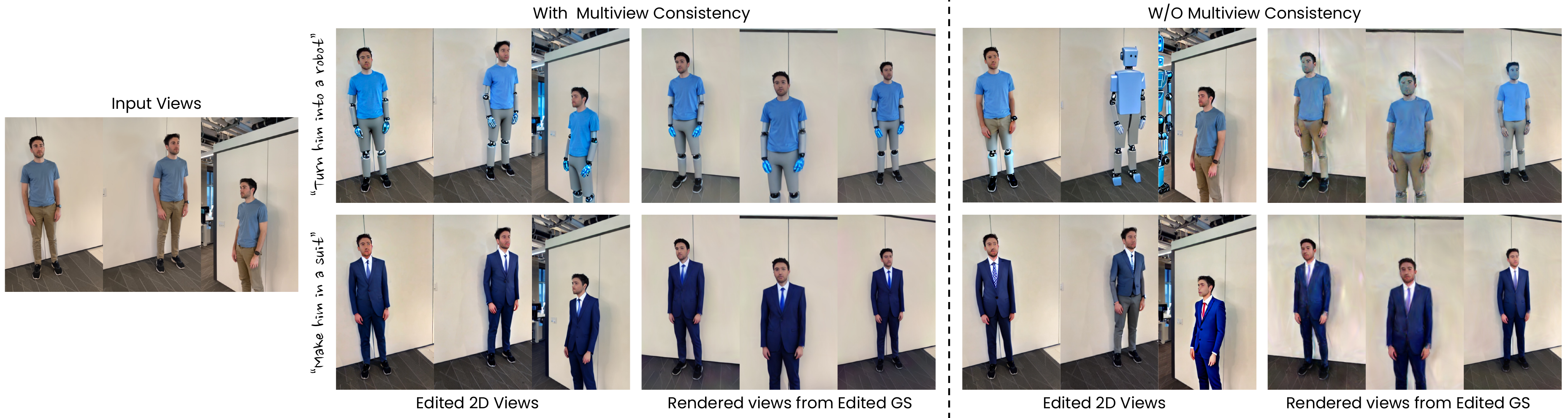}
    \caption{The comparison between with and without multi-view consistency. With the proposed multi-view consistent editing, the edited 3D GS is clear and clean, while without it, it either fails to converge or leads to blurry results.}%
    \label{fig:multiview-consitency}
\end{figure}

In \Cref{fig:main_comparison} we demonstrate visual comparisons with the above methods.
Our proposed method achieves realistic and detailed 3D edits.
One notable advantage of our approach is that ensuring the consistency of the edits in image space results in 3D edits that are overall more faithful to the editing instruction.
In contrast, IDU and SDS work by progressively aligning the edited views of the 3D object, averaging the inconsistencies from the edited images and leading to blurrier or lower-fidelity reconstructions and artifacts.
For example, for the edit instruction \textit{``Make him wear a Venetian mask''}, the texture of the mask generated by \method is detailed and vivid, while other methods fail to properly generate the texture.
ViCA-NeRF results in blurry edits or even fails to produce meaningful edits given prompts such as ``\textit{Turn him into spider man with a mask}''.
It is worth noting that IN2N and ViCA-NeRF generally yield smoother results than GaussianEditor and IP2P+SDS\@; this is due to the fact the NeRF is continuous, while GS is not.

\subsection{Ablation Study}

Next, we ablate the effectiveness of two main components in our editing pipeline: the multi-view consistent editing and the epipolar correspondence matching.
Finally, we demonstrate the convergence speed (in terms of editing) of our approach through a qualitative comparison with \cite{chen2023gaussianeditor}.

\begin{figure}[t]
    \centering
    \includegraphics[width=0.99\linewidth]{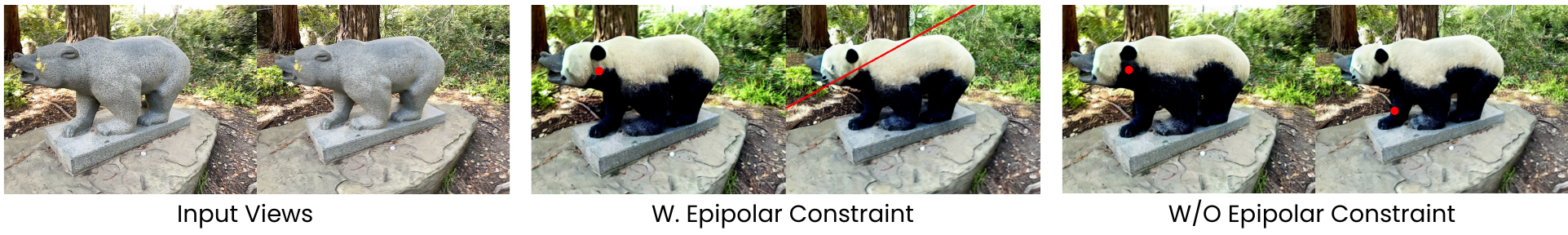}
    \caption{The comparison between edited 2D images with and without epipolar constraints. The one with epipolar constraints successfully matches the correspondences, while the other fails, thus resulting in inconsistent multi-view edits.}%
    \label{fig:epipolar}
\end{figure}

\begin{figure}[t]
    \centering
    \includegraphics[width=0.95\linewidth]{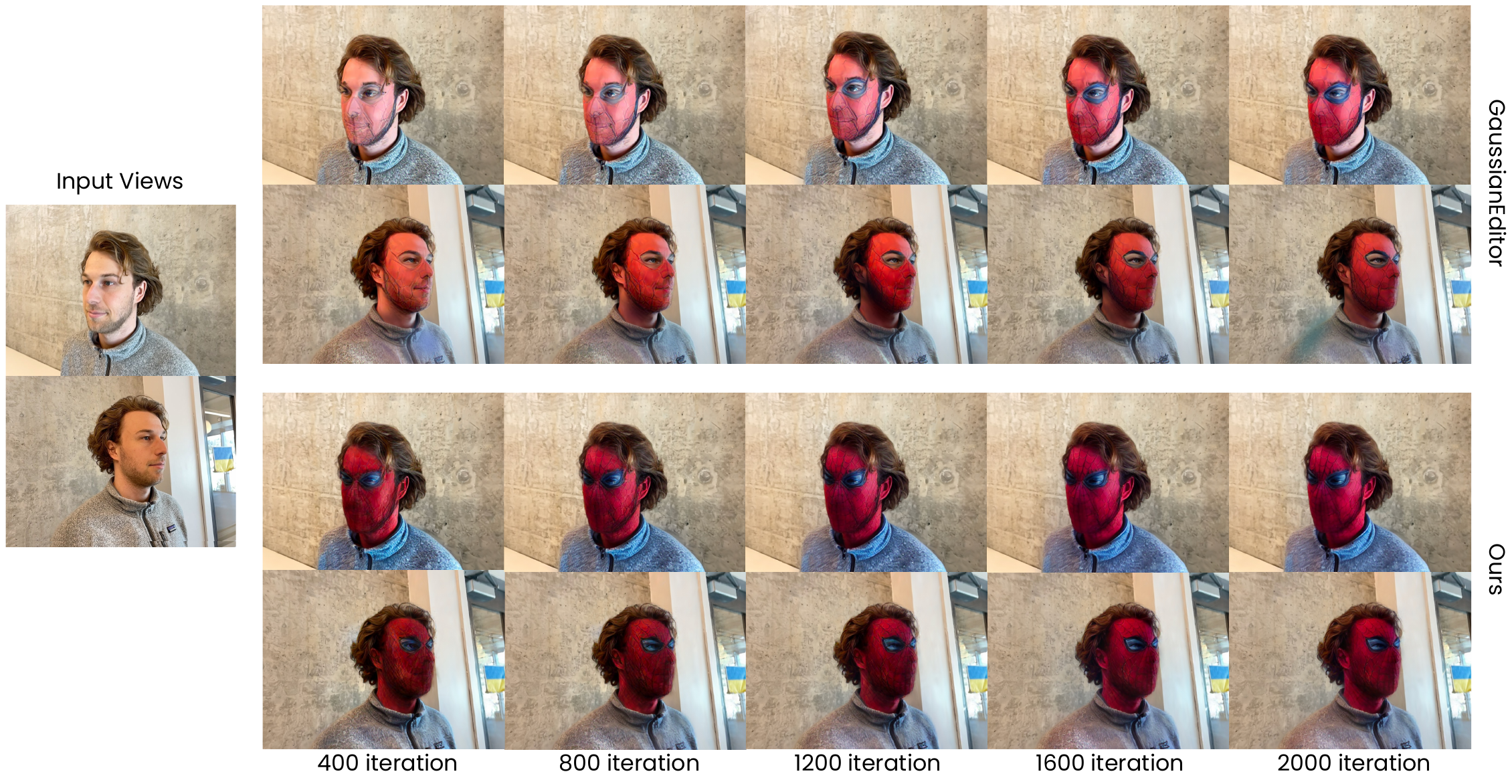}
    \caption{Comparison between our \method and GaussianEditor~\cite{chen2023gaussianeditor} in terms of the number of iterations. Our method achieves realistic editing results with much fewer iterations. With more iterations, our method also gradually refines the details.}%
    \label{fig:effect_of_iteration}
\end{figure}

\begin{figure}[t]
    \centering
    \includegraphics[width=0.95\linewidth]{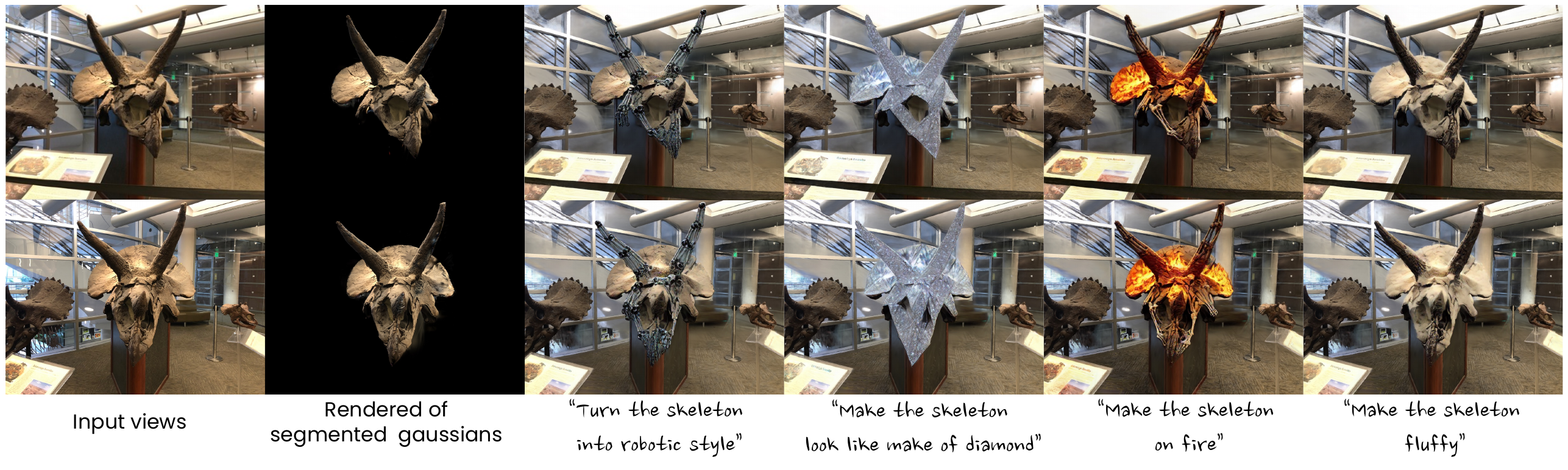}
    \caption{Partial editing results on the \textit{Horns} scene. We achieve realistic editing results only on the required object while keeping the rest unchanged.}%
    \label{fig:partial_editing}
\end{figure}

\subsubsection{Editing without Multi-View Consistency.}

To demonstrate the effectiveness of the multi-view consistent editing, we replace the multi-view consistent editing, by \emph{independently} editing the same views and use those to fit $\G'$.
\Cref{table:ablation} shows the importance of multi-view consistency edits, achieving higher CLIP and CLIP directional scores.
To further demonstrate this, in \cref{fig:multiview-consitency} we show some intermediate 2D editing results with and without multi-view consistent editing.
We observe that multi-view consistent editing produces views with similar appearances, but views can be quite different without it.
This is especially true in the first example (\textit{``turn the man into a robot''}), where the prompt allows for greater variation among views and may even fail to produce sensible edits for some of the frames (the third in this example).
As expected, fitting $\G'$ to these edited views produces a low-quality 3D edited model.
Our multi-view consistent approach clearly alleviates these issues.
We show more detailed comparisons in supplementary material.

\subsubsection{Effect of the Epipolar Constraints.}

In \Cref{table:ablation}, we show that using the epipolar constraints during feature injection (\ie, from key views to other views), both the CLIP score and CLIP directional score increase.
Quantitatively, the impact of using the epipolar constraint appears to be relatively small.
However, we conjecture that this is mainly because the CLIP score assesses the image at a high level while the epipolar constraints help to improve the finer details in editing, such as textures.
The importance of the epipolar constraint is more evident qualitatively. 
In \Cref{fig:epipolar}, the resulting edits without the epipolar constraint are inconsistent between the two views\,---\,red points (just below the bear's ear).
The reason is that it wrongly matches the point with another point in black which leads to failed consistent editing.
The red line in the images indicates the epipolar line corresponding to the red point on the key view.

\subsubsection{Effect of the Number of Iterations.}

We additionally show a comparison between our method and GaussianEditor in \Cref{fig:effect_of_iteration}.
The figure demonstrates that our method achieves high-quality edits after only 400 iterations, while GaussianEditor requires much longer for convergence.
In particular, the results of the 400-iterations column do not use refinement and are generated in \emph{under two minutes}.
With more iterations, the details of the mask in the rendered views are gradually refined. %

\subsection{Partial Editing}

Finally, in \Cref{fig:partial_editing}, we showcase partial (local) editing results on the \textit{Horns} scene from the LLFF~\cite{mildenhall2019llff} dataset.
First, we segment 2D renders to get 2D masks of the skeleton we want to edit. Then, we unproject the 2D masks to 3D getting the mask of the Gaussians.
As last, we render images from different views and edit those views with our proposed \method.
The results are realistic and only applied to the segmented skeleton, demonstrating our method's ability of partial editing.
We provide additional partial editing results in the supplementary material.

\section{Conclusion}%
\label{sec:conclusion}

We have presented \method, a robust framework for \emph{directly} editing of a 3D model by simply fitting it to a small set of views, which are edited by a text-based image editor.
The key insight of our approach is that one must obtain multi-view \emph{consistent} edits in image space, to be able to update the 3D model directly.
Thus we propose an editing mechanism that jointly considers multiple frames based on both appearance cues (image features) and 3D cues (epipolar constraints based on scene geometry).
This approach enables 3D editing that is both more efficient and more faithful to the text instruction in comparison to SDS-like alternatives. 

\subsubsection{Ethics.}

For further details on ethics, data protection, and copyright please see \url{https://www.robots.ox.ac.uk/~vedaldi/research/union/ethics.html}.

\subsubsection{Acknowledgements.}

This research is supported by ERC-CoG UNION 101001212.
I.~L.~is also partially supported by the VisualAI EPSRC grant (EP/T028572/1).

\bibliographystyle{splncs04}
\bibliography{reference,more}

\begin{thebibliography}{100}
\providecommand{\url}[1]{\texttt{#1}}
\providecommand{\urlprefix}{URL }
\providecommand{\doi}[1]{https://doi.org/#1}

\bibitem{bao23sine:}
Bao, C., Zhang, Y., Yang, B., Fan, T., Yang, Z., Bao, H., Zhang, G., Cui, Z.: {SINE:} semantic-driven image-based {NeRF} editing with prior-guided editing field. In: CVPR (2023)

\bibitem{bar2022text2live}
Bar-Tal, O., Ofri-Amar, D., Fridman, R., Kasten, Y., Dekel, T.: Text2live: Text-driven layered image and video editing. In: ECCV (2022)

\bibitem{barron22mip-nerf}
Barron, J.T., Mildenhall, B., Verbin, D., Srinivasan, P.P., Hedman, P.: {Mip-NeRF} 360: Unbounded anti-aliased neural radiance fields. In: CVPR (2022)

\bibitem{brooks2022instructpix2pix}
Brooks, T., Holynski, A., Efros, A.A.: Instructpix2pix: Learning to follow image editing instructions. In: CVPR (2023)

\bibitem{cen2023segment}
Cen, J., Fang, J., Yang, C., Xie, L., Zhang, X., Shen, W., Tian, Q.: Segment any 3d gaussians. arXiv preprint arXiv:2312.00860  (2023)

\bibitem{ceylan2023pix2video}
Ceylan, D., Huang, C.H.P., Mitra, N.J.: Pix2video: Video editing using image diffusion. In: ICCV. pp. 23206--23217 (2023)

\bibitem{chan22efficient}
Chan, E.R., Lin, C.Z., Chan, M.A., Nagano, K., Pan, B., Mello, S.D., Gallo, O., Guibas, L.J., Tremblay, J., Khamis, S., Karras, T., Wetzstein, G.: Efficient geometry-aware {3D} generative adversarial networks. In: CVPR (2022)

\bibitem{chefer2023attendandexcite}
Chefer, H., Alaluf, Y., Vinker, Y., Wolf, L., Cohen-Or, D.: Attend-and-excite: Attention-based semantic guidance for text-to-image diffusion models. In: SIGGRAPH (2023)

\bibitem{chen22tensorf:}
Chen, A., Xu, Z., Geiger, A., Yu, J., Su, H.: {TensoRF}: Tensorial radiance fields (2022)

\bibitem{chen23neuraleditor:}
Chen, J., Lyu, J., Wang, Y.: {NeuralEditor}: Editing neural radiance fields via manipulating point clouds. In: CVPR (2023)

\bibitem{chen2023trainingfree}
Chen, M., Laina, I., Vedaldi, A.: Training-free layout control with cross-attention guidance. In: WACV (2023)

\bibitem{chen24shap-editor}
Chen, M., Xie, J., Laina, I., Vedaldi, A.: {SHAP-EDITOR}: Instruction-guided latent {3D} editing in seconds. In: CVPR (2024)

\bibitem{chen2023gaussianeditor}
Chen, Y., Chen, Z., Zhang, C., Wang, F., Yang, X., Wang, Y., Cai, Z., Yang, L., Liu, H., Lin, G.: Gaussianeditor: Swift and controllable 3d editing with gaussian splatting. CVPR  (2024)

\bibitem{cheng2023progressive3d}
Cheng, X., Yang, T., Wang, J., Li, Y., Zhang, L., Zhang, J., Yuan, L.: Progressive3d: Progressively local editing for text-to-3d content creation with complex semantic prompts. In: ICLR (2024)

\bibitem{chiang22stylizing}
Chiang, P.Z., Tsai, M.S., Tseng, H.Y., sheng Lai, W., Chiu, W.C.: Stylizing {3D} scene via implicit representation and hypernetwork  \textbf{2105.13016} (2022)

\bibitem{dong24vica-nerf:}
Dong, J., Wang, Y.: {ViCA-NeRF}: View-consistency-aware 3d editing of neural radiance fields. In: NeurIPS (2024)

\bibitem{epstein2023selfguidance}
Epstein, D., Jabri, A., Poole, B., Efros, A.A., Holynski, A.: Diffusion self-guidance for controllable image generation. In: NeurIPS (2023)

\bibitem{fang23gaussianeditor:}
Fang, J., Wang, J., Zhang, X., Xie, L., Tian, Q.: Gaussianeditor: Editing 3d gaussians delicately with text instructions. In: CVPR (2024)

\bibitem{gal2022textual}
Gal, R., Alaluf, Y., Atzmon, Y., Patashnik, O., Bermano, A.H., Chechik, G., Cohen-Or, D.: An image is worth one word: Personalizing text-to-image generation using textual inversion. In: ICLR (2023)

\bibitem{gao23textdeformer:}
Gao, W., Aigerman, N., Groueix, T., Kim, V., Hanocka, R.: {TextDeformer}: Geometry manipulation using text guidance. In: SIGGRAPH (2023)

\bibitem{tokenflow2023}
Geyer, M., Bar-Tal, O., Bagon, S., Dekel, T.: Tokenflow: Consistent diffusion features for consistent video editing. In: ICLR (2024)

\bibitem{gong2023recolornerf}
Gong, B., Wang, Y., Han, X., Dou, Q.: Recolornerf: Layer decomposed radiance field for efficient color editing of 3d scenes. In: ACM MM (2023)

\bibitem{gordon2023blended}
Gordon, O., Avrahami, O., Lischinski, D.: Blended-nerf: Zero-shot object generation and blending in existing neural radiance fields. ICCV  (2023)

\bibitem{guler18densepose:}
G{\"{u}}ler, R.A., Neverova, N., Kokkinos, I.: {DensePose}: Dense human pose estimation in the wild. In: CVPR (2018)

\bibitem{haque23instruct-nerf2nerf:}
Haque, A., Tancik, M., Efros, A.A., Holynski, A., Kanazawa, A.: {Instruct-NeRF2NeRF}: Editing {3D} scenes with instructions. In: ICCV (2023)

\bibitem{hartley00multiple}
Hartley, R., Zisserman, A.: Multiple View Geometry in Computer Vision (2000)

\bibitem{hertz2023prompttoprompt}
Hertz, A., Mokady, R., Tenenbaum, J., Aberman, K., Pritch, Y., Cohen-or, D.: Prompt-to-prompt image editing with cross-attention control. In: ICLR (2023)

\bibitem{ho2020denoising}
Ho, J., Jain, A., Abbeel, P.: Denoising diffusion probabilistic models. NeurIPS  \textbf{33},  6840--6851 (2020)

\bibitem{hong22avatarclip:}
Hong, F., Zhang, M., Pan, L., Cai, Z., Yang, L., Liu, Z.: {AvatarCLIP}: zero-shot text-driven generation and animation of {3D} avatars. SIGGRAPH  \textbf{41}(4) (2022)

\bibitem{huang21learning}
Huang, H., Tseng, H., Saini, S., Singh, M., Yang, M.: Learning to stylize novel views. In: ICCV (2021)

\bibitem{huang22stylizednerf:}
Huang, Y., He, Y., Yuan, Y., Lai, Y., Gao, L.: {StylizedNeRF}: Consistent {3D} scene stylization as stylized {NeRF} via {2D-3D} mutual learning. In: CVPR (2022)

\bibitem{huang24seald-nerf:}
Huang, Z., Shi, Y., Bruce, N., Gong, M.: {SealD-NeRF}: Interactive pixel-level editing for dynamic scenes by neural radiance fields  \textbf{2402.13510} (2024)

\bibitem{jambon23nerfshop:}
Jambon, C., Kerbl, B., Kopanas, G., Diolatzis, S., Leimk{\"{u}}hler, T., Drettakis, G.: {NeRFshop:} interactive editing of neural radiance fields. Proc. {ACM} Comput. Graph. Interact. Tech.  \textbf{6}(1) (2023)

\bibitem{kamata2023instruct}
Kamata, H., Sakuma, Y., Hayakawa, A., Ishii, M., Narihira, T.: Instruct 3d-to-3d: Text instruction guided 3d-to-3d conversion. arXiv preprint arXiv:2303.15780  (2023)

\bibitem{kania2022conerf}
Kania, K., Yi, K.M., Kowalski, M., Trzci{\'n}ski, T., Tagliasacchi, A.: Conerf: Controllable neural radiance fields. In: CVPR. pp. 18623--18632 (2022)

\bibitem{kawar2023imagic}
Kawar, B., Zada, S., Lang, O., Tov, O., Chang, H., Dekel, T., Mosseri, I., Irani, M.: Imagic: Text-based real image editing with diffusion models. In: CVPR (2023)

\bibitem{kerbl233d-gaussian}
Kerbl, B., Kopanas, G., Leimk{\"u}hler, T., Drettakis, G.: {3D} {Gaussian Splatting} for real-time radiance field rendering. SIGGRAPH  \textbf{42}(4) (2023)

\bibitem{khachatryan2023text2video}
Khachatryan, L., Movsisyan, A., Tadevosyan, V., Henschel, R., Wang, Z., Navasardyan, S., Shi, H.: Text2video-zero: Text-to-image diffusion models are zero-shot video generators. In: ICCV (2023)

\bibitem{kirillov23segment}
Kirillov, A., Mintun, E., Ravi, N., Mao, H., Rolland, C., Gustafson, L., Xiao, T., Whitehead, S., Berg, A.C., Lo, W.Y., Doll{\'a}r, P., Girshick, R.: Segment anything. In: CVPR (2023)

\bibitem{kobayashi2022decomposing}
Kobayashi, S., Matsumoto, E., Sitzmann, V.: Decomposing nerf for editing via feature field distillation. NeurIPS  \textbf{35},  23311--23330 (2022)

\bibitem{kuang23palettenerf:}
Kuang, Z., Luan, F., Bi, S., Shu, Z., Wetzstein, G., Sunkavalli, K.: {PaletteNeRF}: Palette-based appearance editing of neural radiance fields. In: CVPR (2023)

\bibitem{kumari2022customdiffusion}
Kumari, N., Zhang, B., Zhang, R., Shechtman, E., Zhu, J.Y.: Multi-concept customization of text-to-image diffusion. In: CVPR (2023)

\bibitem{lazova23control-nerf:}
Lazova, V., Guzov, V., Olszewski, K., Tulyakov, S., Pons{-}Moll, G.: {Control-NeRF}: Editable feature volumes for scene rendering and manipulation. In: WACV (2023)

\bibitem{lee2023ice}
Lee, J.H., Kim, D.S.: Ice-nerf: Interactive color editing of nerfs via decomposition-aware weight optimization. In: ICCV. pp. 3491--3501 (2023)

\bibitem{lei2022tango}
Lei, J., Zhang, Y., Jia, K., et~al.: Tango: Text-driven photorealistic and robust 3d stylization via lighting decomposition. NeurIPS  \textbf{35},  30923--30936 (2022)

\bibitem{li2022languagedriven}
Li, B., Weinberger, K.Q., Belongie, S., Koltun, V., Ranftl, R.: Language-driven semantic segmentation. In: ICLR (2022)

\bibitem{li223ddesigner:}
Li, G., Zheng, H., Wang, C., Li, C., Zheng, C., Tao, D.: {3DDesigner}: Towards photorealistic 3d object generation and editing with text-guided diffusion models  \textbf{abs/2211.14108} (2022)

\bibitem{li2022grounded}
Li, L.H., Zhang, P., Zhang, H., Yang, J., Li, C., Zhong, Y., Wang, L., Yuan, L., Zhang, L., Hwang, J.N., et~al.: Grounded language-image pre-training. In: CVPR. pp. 10965--10975 (2022)

\bibitem{li2023focaldreamer}
Li, Y., Dou, Y., Shi, Y., Lei, Y., Chen, X., Zhang, Y., Zhou, P., Ni, B.: Focaldreamer: Text-driven 3d editing via focal-fusion assembly. In: AAAI (2024)

\bibitem{li2023gligen}
Li, Y., Liu, H., Wu, Q., Mu, F., Yang, J., Gao, J., Li, C., Lee, Y.J.: Gligen: Open-set grounded text-to-image generation. In: CVPR (2023)

\bibitem{lin23componerf:}
Lin, Y., Bai, H., Li, S., Lu, H., Lin, X., Xiong, H., Wang, L.: {CompoNeRF}: Text-guided multi-object compositional nerf with editable 3d scene layout  \textbf{abs/2303.13843} (2023)

\bibitem{liu22nerf-in:}
Liu, H., Shen, I., Chen, B.: {NeRF-In}: Free-form {NeRF} inpainting with {RGB-D} priors  \textbf{abs/2206.04901} (2022)

\bibitem{liu21editing}
Liu, S., Zhang, X., Zhang, Z., Zhang, R., Zhu, J., Russell, B.: Editing conditional radiance fields. In: ICCV (2021)

\bibitem{mascaro21diffuser:}
Mascaro, R., Teixeira, L., Chli, M.: Diffuser: Multi-view {2D}-to-{3D} label diffusion for semantic scene segmentation (2021)

\bibitem{melas-kyriazi24im-3d}
Melas-Kyriazi, L., Laina, I., Rupprecht, C., Neverova, N., Vedaldi, A., Gafni, O., Kokkinos, F.: {IM-3D}: Iterative multiview diffusion and reconstruction for high-quality {3D} generation. In: ICML (2024)

\bibitem{melas-kyriazi23realfusion}
Melas{-}Kyriazi, L., Rupprecht, C., Laina, I., Vedaldi, A.: {RealFusion}: 360 reconstruction of any object from a single image. In: CVPR (2023)

\bibitem{meng2022sdedit}
Meng, C., He, Y., Song, Y., Song, J., Wu, J., Zhu, J.Y., Ermon, S.: {SDE}dit: Guided image synthesis and editing with stochastic differential equations. In: ICLR (2022)

\bibitem{michel22text2mesh:}
Michel, O., Bar{-}On, R., Liu, R., Benaim, S., Hanocka, R.: {Text2Mesh:} text-driven neural stylization for meshes. In: CVPR (2022)

\bibitem{mikaeili23sked:}
Mikaeili, A., Perel, O., Safaee, M., Cohen{-}Or, D., Mahdavi{-}Amiri, A.: {SKED:} sketch-guided text-based {3D} editing. In: ICCV (2023)

\bibitem{mildenhall2019llff}
Mildenhall, B., Srinivasan, P.P., Ortiz-Cayon, R., Kalantari, N.K., Ramamoorthi, R., Ng, R., Kar, A.: Local light field fusion: Practical view synthesis with prescriptive sampling guidelines. ACM Transactions on Graphics (TOG)  (2019)

\bibitem{mildenhall20nerf:}
Mildenhall, B., Srinivasan, P.P., Tancik, M., Barron, J.T., Ramamoorthi, R., Ng, R.: {NeRF}: Representing scenes as neural radiance fields for view synthesis. In: ECCV (2020)

\bibitem{mirzaei23watch}
Mirzaei, A., Aumentado{-}Armstrong, T., Brubaker, M.A., Kelly, J., Levinshtein, A., Derpanis, K.G., Gilitschenski, I.: Watch your steps: Local image and scene editing by text instructions  \textbf{abs/2308.08947} (2023)

\bibitem{mirzaei2023spin}
Mirzaei, A., Aumentado-Armstrong, T., Derpanis, K.G., Kelly, J., Brubaker, M.A., Gilitschenski, I., Levinshtein, A.: Spin-nerf: Multiview segmentation and perceptual inpainting with neural radiance fields. In: CVPR. pp. 20669--20679 (2023)

\bibitem{mokady2023null}
Mokady, R., Hertz, A., Aberman, K., Pritch, Y., Cohen-Or, D.: Null-text inversion for editing real images using guided diffusion models. In: CVPR. pp. 6038--6047 (2023)

\bibitem{nguyen-phuoc22snerf:}
Nguyen{-}Phuoc, T., Liu, F., Xiao, L.: {SNeRF}: stylized neural implicit representations for 3d scenes. SIGGRAPH  \textbf{41}(4) (2022)

\bibitem{nichol21glide:}
Nichol, A., Dhariwal, P., Ramesh, A., Shyam, P., Mishkin, P., McGrew, B., Sutskever, I., Chen, M.: {GLIDE:} towards photorealistic image generation and editing with text-guided diffusion models  \textbf{abs/2112.10741} (2021)

\bibitem{park2023ed}
Park, J., Kwon, G., Ye, J.C.: Ed-nerf: Efficient text-guided editing of 3d scene using latent space nerf. In: ICLR (2024)

\bibitem{parmar2023zeroshot}
Parmar, G., Singh, K.K., Zhang, R., Li, Y., Lu, J., Zhu, J.Y.: Zero-shot image-to-image translation. In: SIGGRAPH (2023)

\bibitem{podell23sdxl:}
Podell, D., English, Z., Lacey, K., Blattmann, A., Dockhorn, T., M{\"{u}}ller, J., Penna, J., Rombach, R.: {SDXL:} improving latent diffusion models for high-resolution image synthesis  \textbf{abs/2307.01952} (2023)

\bibitem{poole2022dreamfusion}
Poole, B., Jain, A., Barron, J.T., Mildenhall, B.: Dreamfusion: Text-to-3d using 2d diffusion. In: ICLR (2023)

\bibitem{qi2023fatezero}
Qi, C., Cun, X., Zhang, Y., Lei, C., Wang, X., Shan, Y., Chen, Q.: Fatezero: Fusing attentions for zero-shot text-based video editing. ICCV  (2023)

\bibitem{radford21learning}
Radford, A., Kim, J.W., Hallacy, C., Ramesh, A., Goh, G., Agarwal, S., Sastry, G., Askell, A., Mishkin, P., Clark, J., Krueger, G., Sutskever, I.: Learning transferable visual models from natural language supervision. In: ICML. vol.~139, pp. 8748--8763 (2021)

\bibitem{rombach22high-resolution}
Rombach, R., Blattmann, A., Lorenz, D., Esser, P., Ommer, B.: High-resolution image synthesis with latent diffusion models. In: CVPR (2022)

\bibitem{ruiz2023dreambooth}
Ruiz, N., Li, Y., Jampani, V., Pritch, Y., Rubinstein, M., Aberman, K.: Dreambooth: Fine tuning text-to-image diffusion models for subject-driven generation. In: CVPR (2023)

\bibitem{sella23vox-e:}
Sella, E., Fiebelman, G., Hedman, P., Averbuch{-}Elor, H.: Vox-e: Text-guided voxel editing of 3d objects. In: ICCV (2023)

\bibitem{shi23dragdiffusion:}
Shi, Y., Xue, C., Pan, J., Zhang, W., Tan, V.Y., Bai, S.: {DragDiffusion}: Harnessing diffusion models for interactive point-based image editing  \textbf{2306.14435} (2023)

\bibitem{song2023blending}
Song, H., Choi, S., Do, H., Lee, C., Kim, T.: Blending-nerf: Text-driven localized editing in neural radiance fields. In: ICCV. pp. 14383--14393 (2023)

\bibitem{song21denoising}
Song, J., Meng, C., Ermon, S.: Denoising diffusion implicit models. In: ICLR (2021)

\bibitem{song2021denoising}
Song, J., Meng, C., Ermon, S.: Denoising diffusion implicit models. In: ICLR (2021)

\bibitem{sun22direct}
Sun, C., Sun, M., Chen, H.: Direct voxel grid optimization: Super-fast convergence for radiance fields reconstruction. In: CVPR (2022)

\bibitem{sun22nerfeditor:}
Sun, C., Liu, Y., Han, J., Gould, S.: {NeRFEditor:} differentiable style decomposition for full {3D} scene editing. In: WACV (2022)

\bibitem{teng23drag-a-video:}
Teng, Y., Xie, E., Wu, Y., Han, H., Li, Z., Liu, X.: Drag-a-video: Non-rigid video editing with point-based interaction  \textbf{abs/2312.02936} (2023)

\bibitem{tretschk23scenerflow:}
Tretschk, E., Golyanik, V., Zollh{\"{o}}fer, M., Bozic, A., Lassner, C., Theobalt, C.: Scenerflow: Time-consistent reconstruction of general dynamic scenes  \textbf{abs/2308.08258} (2023)

\bibitem{tschernezki22neural}
Tschernezki, V., Laina, I., Larlus, D., Vedaldi, A.: {Neural Feature Fusion Fields}: {3D} distillation of self-supervised {2D} image representation (2022)

\bibitem{Tumanyan_2023_CVPR}
Tumanyan, N., Geyer, M., Bagon, S., Dekel, T.: Plug-and-play diffusion features for text-driven image-to-image translation. In: CVPR (2023)

\bibitem{vargas23denoising}
Vargas, F., Grathwohl, W.S., Doucet, A.: Denoising diffusion samplers. In: ICLR (2023)

\bibitem{wang23proteusnerf:}
Wang, B., Dutt, N.S., Mitra, N.J.: {ProteusNeRF}: Fast lightweight {NeRF} editing using {3D}-aware image context  \textbf{abs/2310.09965} (2023)

\bibitem{wang22clip-nerf:}
Wang, C., Chai, M., He, M., Chen, D., Liao, J.: {CLIP-NeRF}: Text-and-image driven manipulation of neural radiance fields. In: CVPR (2022)

\bibitem{wang22nerf-art:}
Wang, C., Jiang, R., Chai, M., He, M., Chen, D., Liao, J.: {NeRF-Art}: Text-driven neural radiance fields stylization  \textbf{abs/2212.08070} (2022)

\bibitem{wang2023inpaintnerf360}
Wang, D., Zhang, T., Abboud, A., S{\"u}sstrunk, S.: Inpaintnerf360: Text-guided 3d inpainting on unbounded neural radiance fields. arXiv preprint arXiv:2305.15094  (2023)

\bibitem{wang2024gaussianeditor}
Wang, J., Fang, J., Zhang, X., Xie, L., Tian, Q.: Gaussianeditor: Editing 3d gaussians delicately with text instructions. In: Proceedings of the IEEE/CVF Conference on Computer Vision and Pattern Recognition. pp. 20902--20911 (2024)

\bibitem{wang23seal-3d:}
Wang, X., Zhu, J., Ye, Q., Huo, Y., Ran, Y., Zhong, Z., Chen, J.: {Seal-3D}: Interactive pixel-level editing for neural radiance fields. In: ICCV (2023)

\bibitem{weder2023removing}
Weder, S., Garcia-Hernando, G., Monszpart, A., Pollefeys, M., Brostow, G.J., Firman, M., Vicente, S.: Removing objects from neural radiance fields. In: CVPR. pp. 16528--16538 (2023)

\bibitem{wu2023tune}
Wu, J.Z., Ge, Y., Wang, X., Lei, S.W., Gu, Y., Shi, Y., Hsu, W., Shan, Y., Qie, X., Shou, M.Z.: Tune-a-video: One-shot tuning of image diffusion models for text-to-video generation. In: Proceedings of the IEEE/CVF International Conference on Computer Vision. pp. 7623--7633 (2023)

\bibitem{xu23instructp2p:}
Xu, J., Wang, X., Cao, Y., Cheng, W., Shan, Y., Gao, S.: {InstructP2P}: Learning to edit {3D} point clouds with text instructions  \textbf{abs/2306.07154} (2023)

\bibitem{xu2023desrf}
Xu, S., Li, L., Shen, L., Lian, Z.: Desrf: Deformable stylized radiance field. In: CVPR. pp. 709--718 (2023)

\bibitem{xu22deforming}
Xu, T., Harada, T.: Deforming radiance fields with cages. In: ECCV (2022)

\bibitem{yang22neumesh:}
Yang, B., Bao, C., Zeng, J., Bao, H., Zhang, Y., Cui, Z., Zhang, G.: Neumesh: Learning disentangled neural mesh-based implicit field for geometry and texture editing. In: ECCV (2022)

\bibitem{yang21learning}
Yang, B., Zhang, Y., Xu, Y., Li, Y., Zhou, H., Bao, H., Zhang, G., Cui, Z.: Learning object-compositional neural radiance field for editable scene rendering. In: ICCV (2021)

\bibitem{yu2023edit}
Yu, L., Xiang, W., Han, K.: Edit-diffnerf: Editing 3d neural radiance fields using 2d diffusion model. arXiv preprint arXiv:2306.09551  (2023)

\bibitem{yuan22nerf-editing:}
Yuan, Y., Sun, Y., Lai, Y., Ma, Y., Jia, R., Gao, L.: {NeRF}-editing: Geometry editing of neural radiance fields. In: CVPR (2022)

\bibitem{zhang2023text}
Zhang, H., Feng, Y., Kulits, P., Wen, Y., Thies, J., Black, M.J.: Text-guided generation and editing of compositional 3d avatars. In: 2024 International Conference on 3D Vision (3DV) (2024)

\bibitem{zhang21editable}
Zhang, J., Liu, X., Ye, X., Zhao, F., Zhang, Y., Wu, M., Zhang, Y., Xu, L., Yu, J.: Editable free-viewpoint video using a layered neural representation. {ACM} Trans. Graph.  \textbf{40}(4) (2021)

\bibitem{zhang22arf:}
Zhang, K., Kolkin, N., Bi, S., Luan, F., Xu, Z., Shechtman, E., Snavely, N.: {ARF}: Artistic radiance fields. In: ECCV (2022)

\bibitem{Zhang2023MagicBrush}
Zhang, K., Mo, L., Chen, W., Sun, H., Su, Y.: Magicbrush: A manually annotated dataset for instruction-guided image editing. In: NeurIPS (2023)

\bibitem{Zhang_2023_ICCV}
Zhang, L., Rao, A., Agrawala, M.: Adding conditional control to text-to-image diffusion models. In: ICCV (2023)

\bibitem{zhang2018unreasonable}
Zhang, R., Isola, P., Efros, A.A., Shechtman, E., Wang, O.: The unreasonable effectiveness of deep features as a perceptual metric. In: Proceedings of the IEEE conference on computer vision and pattern recognition. pp. 586--595 (2018)

\bibitem{zhang2023hive}
Zhang, S., Yang, X., Feng, Y., Qin, C., Chen, C.C., Yu, N., Chen, Z., Wang, H., Savarese, S., Ermon, S., Xiong, C., Xu, R.: Hive: Harnessing human feedback for instructional visual editing. In: CVPR (2024)

\bibitem{zheng23editablenerf:}
Zheng, C., Lin, W., Xu, F.: {EditableNeRF}: Editing topologically varying neural radiance fields by key points. In: CVPR (2023)

\bibitem{zhou2024feature}
Zhou, S., Chang, H., Jiang, S., Fan, Z., Zhu, Z., Xu, D., Chari, P., You, S., Wang, Z., Kadambi, A.: Feature 3dgs: Supercharging 3d gaussian splatting to enable distilled feature fields. In: Proceedings of the {IEEE} Conference on Computer Vision and Pattern Recognition ({CVPR}). pp. 21676--21685 (2024)

\bibitem{zhou2023repaint}
Zhou, X., He, Y., Yu, F.R., Li, J., Li, Y.: Repaint-nerf: Nerf editting via semantic masks and diffusion models. In: IJCAI (2023)

\bibitem{zhuang23dreameditor:}
Zhuang, J., Wang, C., Lin, L., Liu, L., Li, G.: {DreamEditor}: Text-driven {3D} scene editing with neural fields. In: SIGGRAPH (2023)

\end{thebibliography}

\section*{Appendix}
\label{appendix}
\appendix

This supplementary material contains the following parts:
\begin{itemize}

    \item \textbf{\nameref*{sec:more_results} (\Cref{sec:more_results}).} We provide additional qualitative examples of our method, including more editing prompts and partial editing results on complex scenes with many objects.

    \item \textbf{\nameref*{sec:additional_comparison} (\Cref{sec:additional_comparison}).} We provide in-depth comparisons of our approach and other update schemes and further comparisons with prior work, GaussianEditor. 

    \item \textbf{\nameref*{sec:addtional_details} (\Cref{sec:addtional_details}).} We provide more details regarding the experimental implementation and evaluation protocol.
    
    \item \textbf{\nameref*{sec:social_limitation} (\Cref{sec:social_limitation}).} We discuss the social impact and limitations of our work.

\end{itemize}

\section{Additional Editing Results} 
\label{sec:more_results}
In this section, we provide additional qualitative editing results of \method with more prompts and additional partial editing results on the Mip-NeRF360 \cite{barron22mip-nerf} dataset.

\subsubsection{Results with Additional Prompts.}

\begin{figure}[t]
    \centering
    \includegraphics[width=\linewidth]{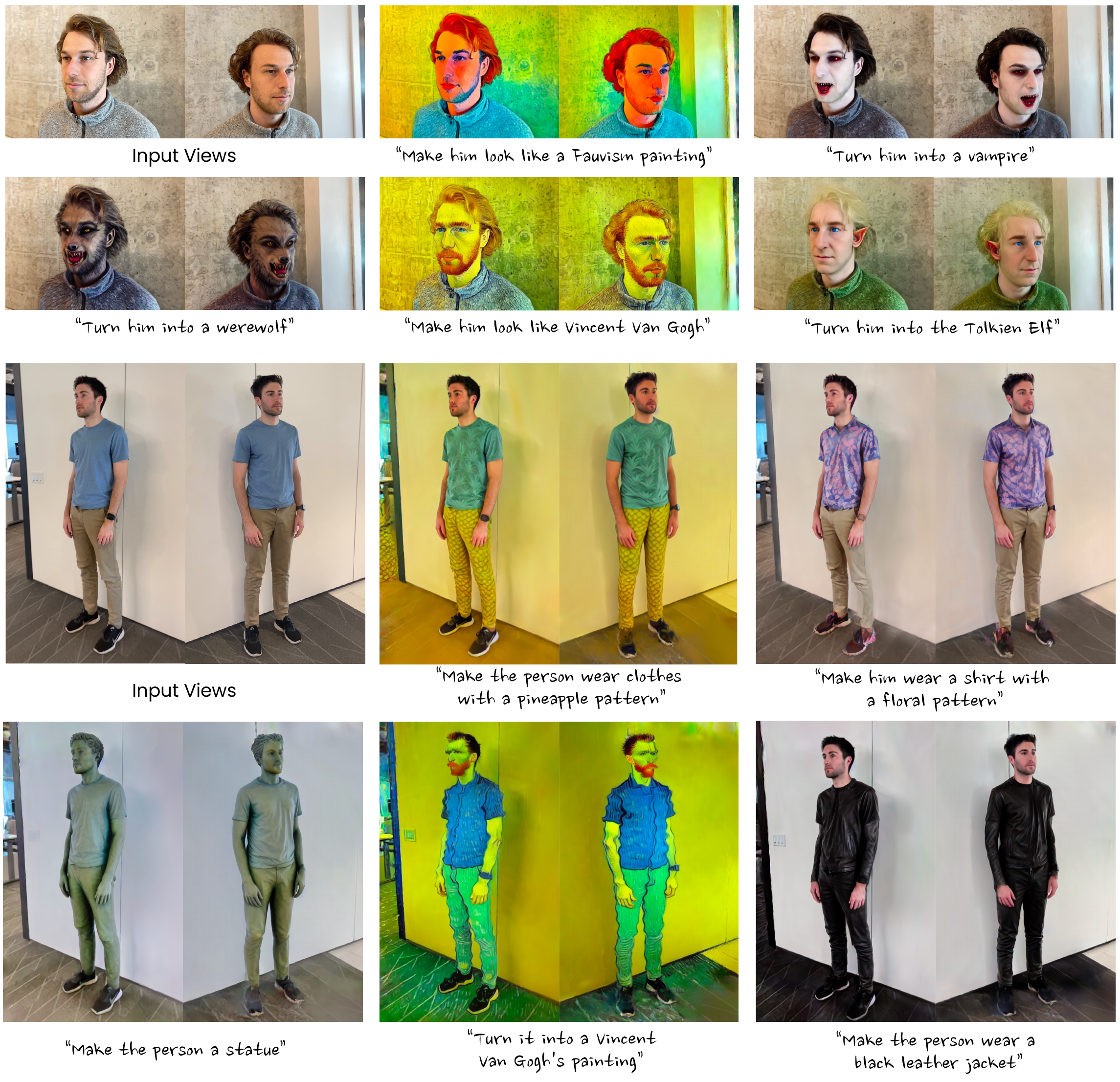}
    \vspace{-1em}
    \caption{\textbf{Additional editing examples.} We provide more qualitative results, including prompts to change the character or style. Our \method produces realistic editing results with detailed textures.}
    \label{fig:more_visual_result}
\end{figure}

In \Cref{fig:more_visual_result}, we provide further qualitative editing results from our method, demonstrating the ability of \method to generate complex and visually compelling editing effects.
Specifically, when tasked with transforming the appearance of a human figure into mythical creatures, such as a ``\textit{Vampire},'' ``\textit{Tolkien Elf},'' or \textit{Werewolf},'' \method produces highly realistic results.
This capability is not solely restricted to character modifications.
We also show the ability of \method to handle style transfer, such as converting the 3D models' style into ``\textit{Fauvism painting}''.
Finally, by directly and consistently editing multiple images at once, we show that \method produces intricate 3D edits with detailed patterns and textures, \eg, changing the clothes of the human figure to a ``\textit{pineapple pattern}'' or ``\textit{floral pattern}''.

\subsubsection{Results with Partial Editing.}
\begin{figure}[t]
    \centering
    \includegraphics[width=\linewidth]{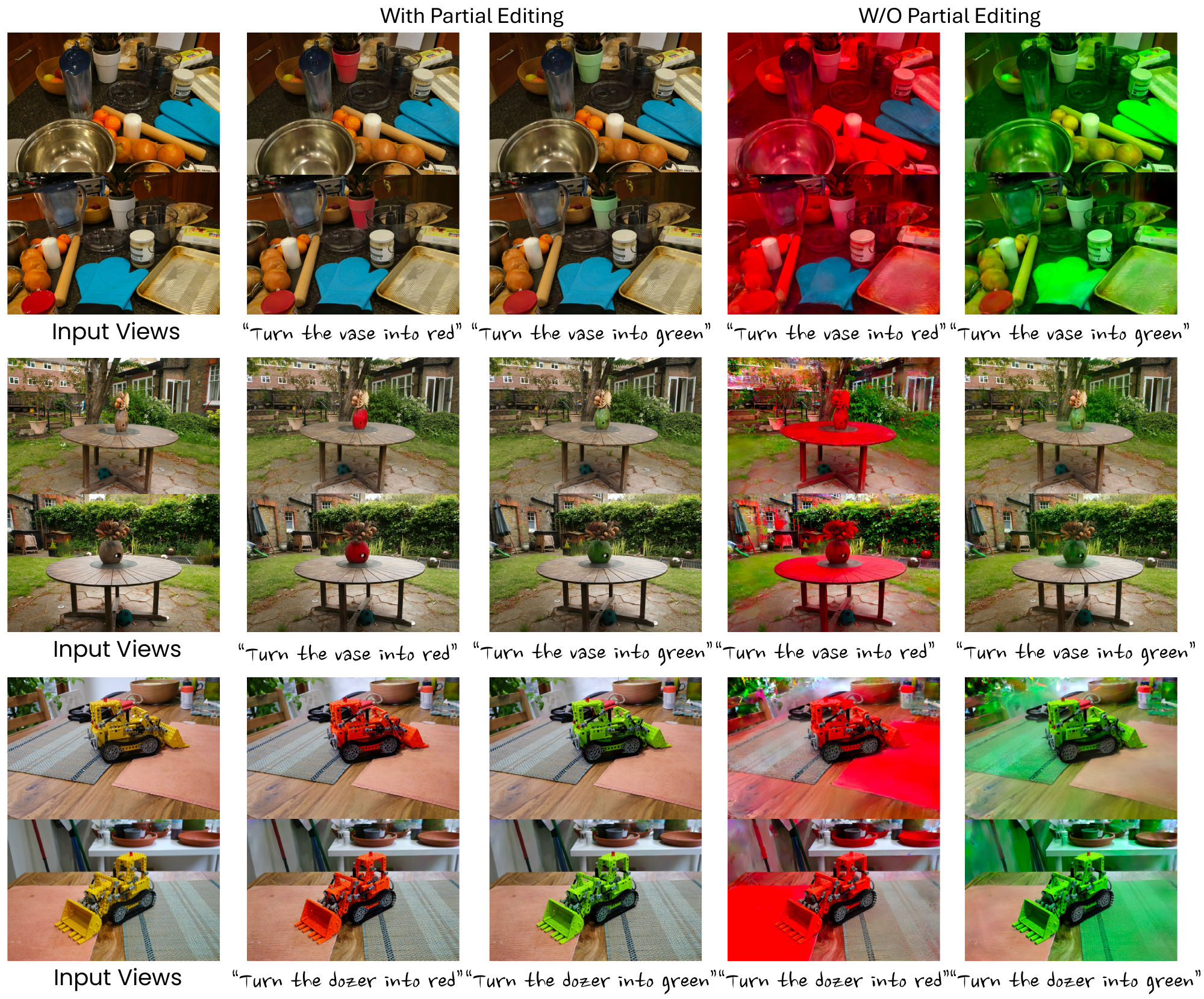}
    \vspace{-0.5em}
    \caption{Additional partial editing results. Our \method successfully changes only the selected objects faithfully. We can even tell the color change through the transparent bowl before the vase.}
    \label{fig:partial_supp}
\end{figure}

We provide additional partial editing results using three different scenes provided by Mip-NeRF360 \cite{barron22mip-nerf} in \Cref{fig:partial_supp}. Our \method successfully edits the color of the selected objects such as \textit{vase} and \textit{dozer} while keeping other stuff unchanged. For example, only the vase in the first row is changed to either green or red. Besides, we can still distinguish the color changes from the transparent bowl before the vase. As shown in the last two columns of the figure, without partial editing, the whole scene turns either red or green, which clearly demonstrates the advantages of partial editing. The core issue is the inability of the image editor (InstructPix2Pix) to isolate and modify the color of just the selected object, affecting the entire image instead.

\begin{figure}[t]
    \centering
    \includegraphics[width=\linewidth]{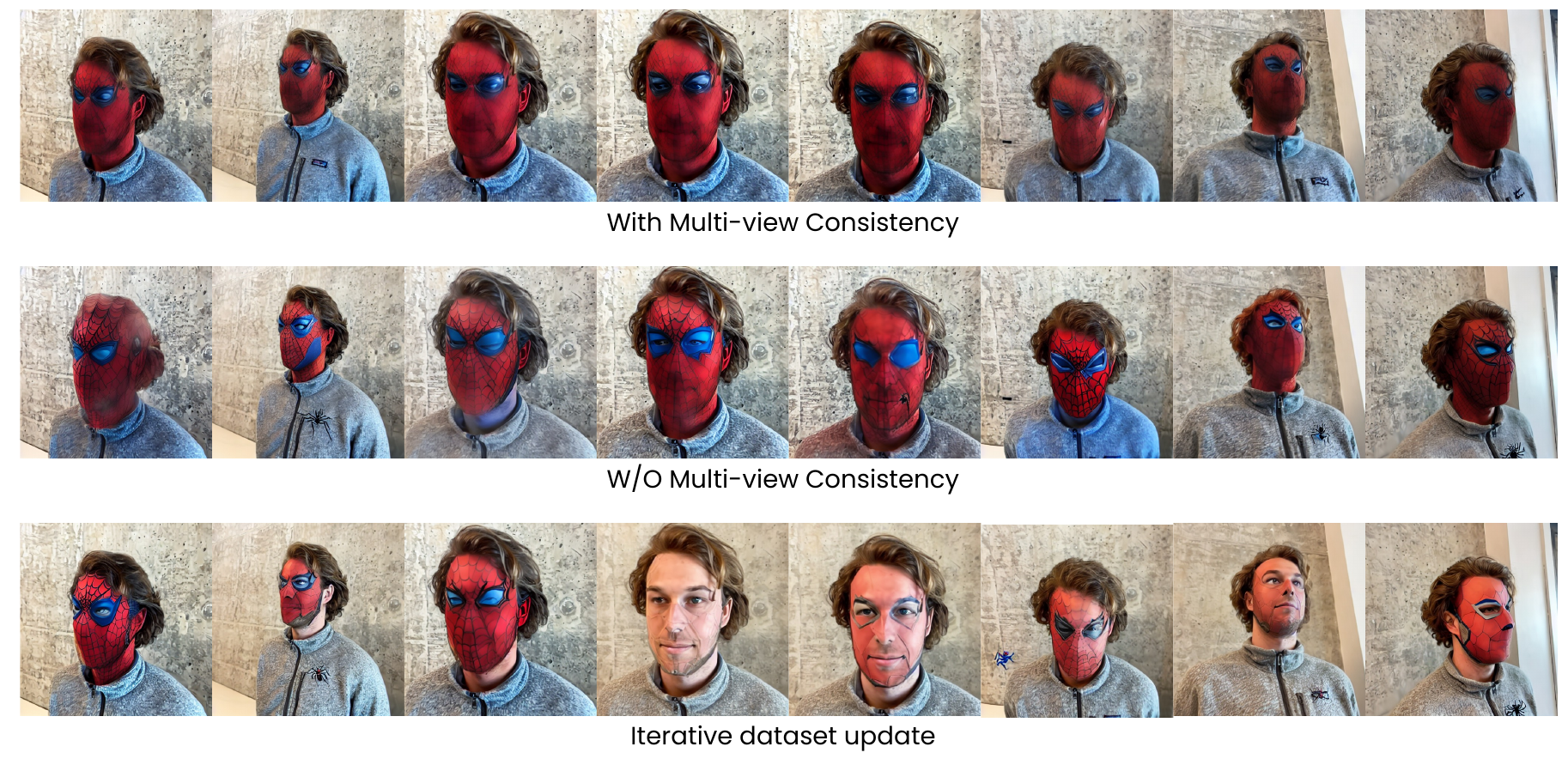}
    \vspace{-1em}
    \caption{\textbf{Comparison of different updating methods.} This figure compares the 2D editing results (i) with multi-view consistency, (ii) without multi-view consistency, and (iii) with iterative dataset updates. Multi-view consistent editing yields a consistent appearance across all different views.}
    \label{fig:multiview_supp}
\end{figure}

\begin{figure}[!t]
  \begin{minipage}[c]{0.5\linewidth}
    \centering
    \includegraphics[width=\linewidth]{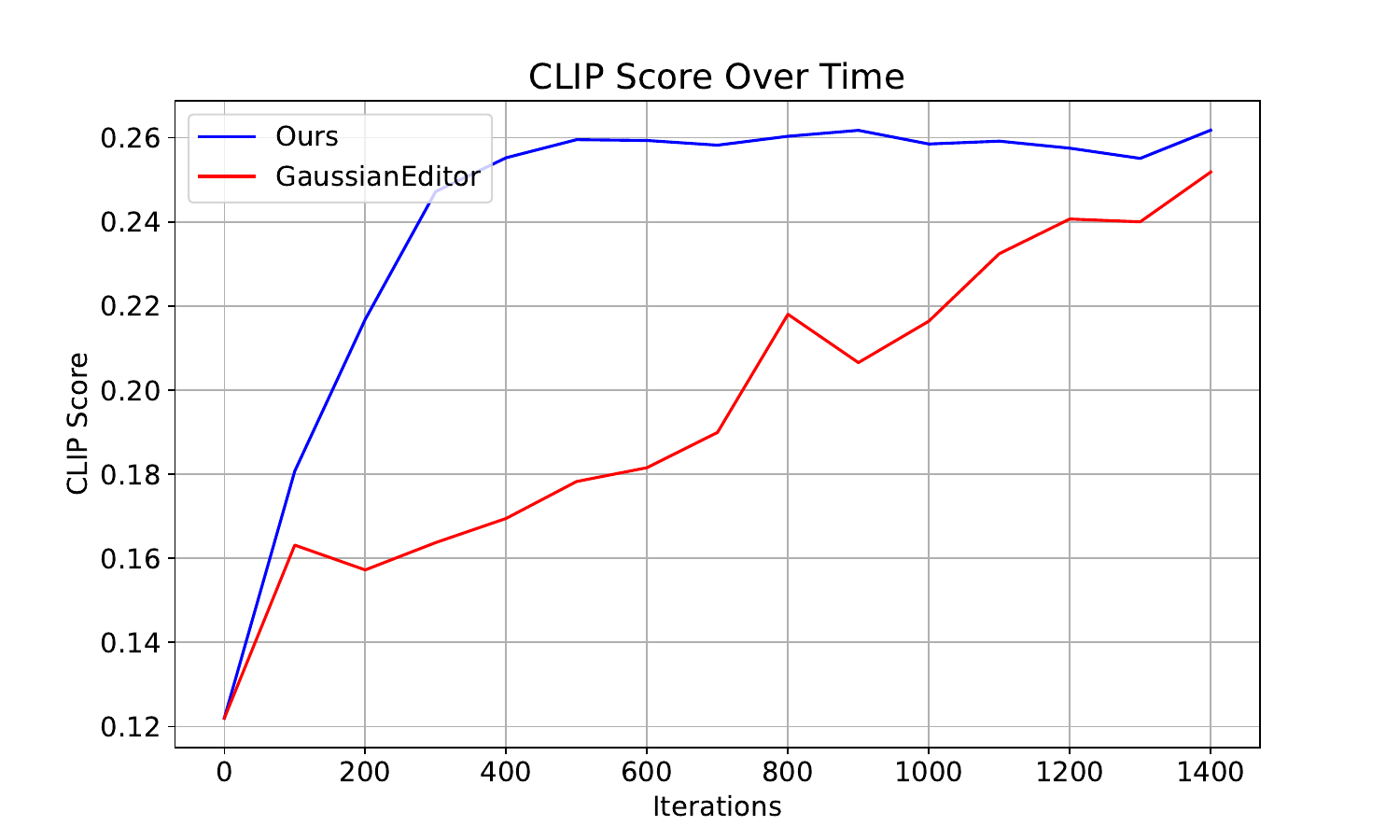}
  \end{minipage}
  \begin{minipage}[c]{0.5\linewidth}
    \centering
    \includegraphics[width=\linewidth]{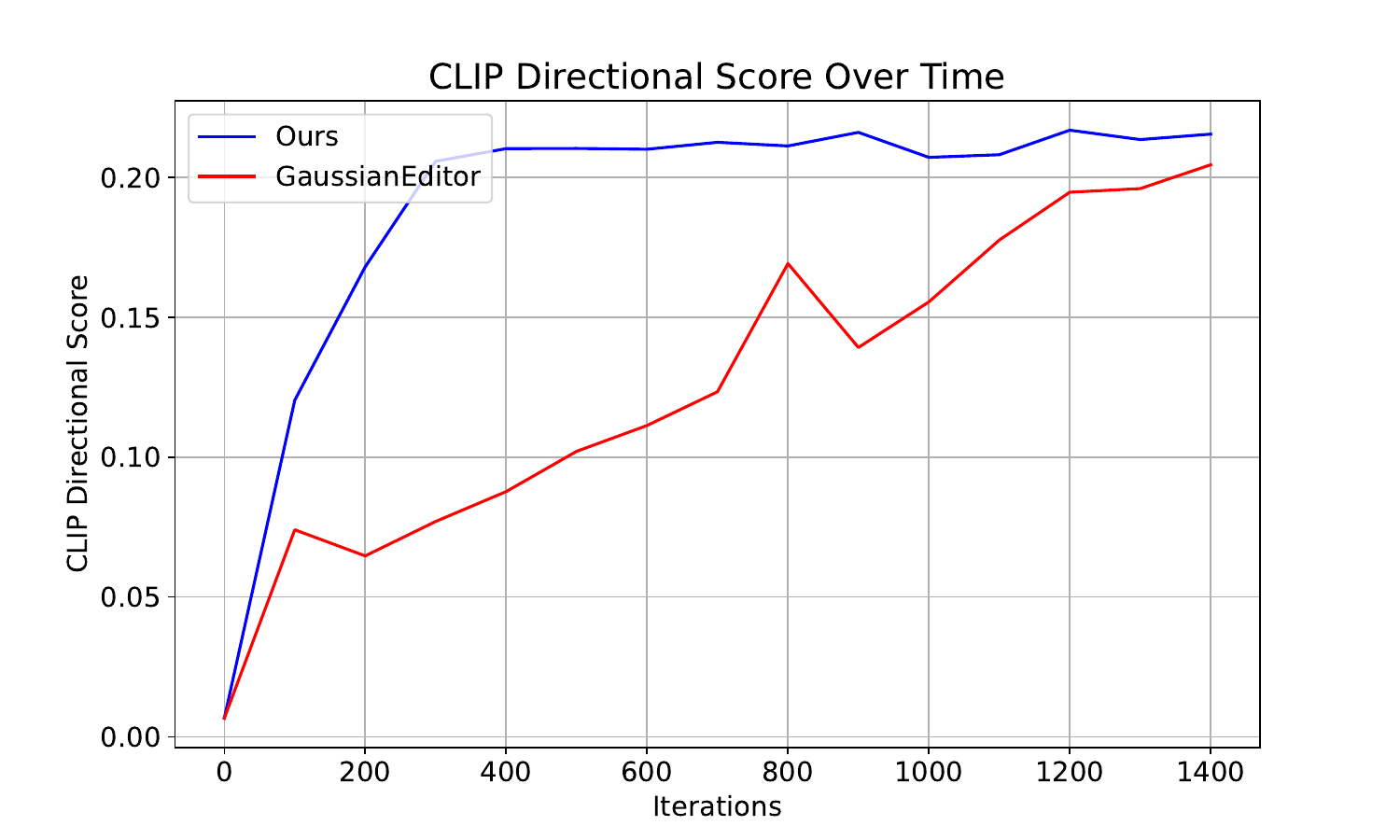}
  \end{minipage}
    \caption{\textbf{CLIP and CLIP directional scores w.r.t. the number of iterations.} We compare the CLIP scores of \method and GaussianEditor over training iterations for the example depicted in \Cref{fig:multiview_supp}. Our method achieves high CLIP and CLIP directional scores in under 400 iterations, which is before the additional refinement step. We perform refining after 500 iterations.}
    \label{fig:plot}
\end{figure}

\section{Additional Comparisons}
\label{sec:additional_comparison}

\subsubsection{Multi-view Consistent Editing vs.~Iterative Dataset Updates.}
Next, we provide a qualitative comparison of our \emph{direct} approach \textit{with} and \textit{without} multi-view consistency, as well as the iterative dataset update scheme.
In \Cref{fig:multiview_supp}, we show the 2D edited images that are later used to update the corresponding 3D model.
Note that all variants use the same base 3D model. 
We observe that when employing multi-view consistency in 2D editing, we achieve a uniform and consistent appearance across all images, which in turn accelerates the reconstruction process.
Conversely, without multi-view consistency, the images are edited independently by InstructPix2Pix.
As a result, the appearance of the edited images tends to vary, despite the input views being similar (notably, the 3rd to 5th images in the middle row).
Finally, the inconsistencies are exacerbated under the iterative dataset update scheme, which involves introducing different amounts of noise, resulting in considerable variability after editing. This variability contributes to the slower convergence speeds of methods such as IN2N \cite{haque23instruct-nerf2nerf:} and GE \cite{chen2023gaussianeditor}.

\subsubsection{Extended Comparisons with GaussianEditor.}
GaussianEditor \cite{chen2023gaussianeditor} is a recent work that attempts to edit Gaussian Splatting-based models similarly to IN2N \cite{haque23instruct-nerf2nerf:}.
Based on the iterative dataset update scheme, both GaussianEditor and IN2N iterate between 3D model updating and image editing.
As discussed above, iterative dataset updates require many iterations to successfully apply the desired editing effect to the 3D model.
In contrast, our \method edits rendered images \emph{simultaneously} and subsequently updates the Gaussian Splatting model, achieving faster speed and better quality.
\Cref{fig:plot} further demonstrates this by comparing the number of iterations required for convergence between \method and GaussianEditor.
By directly integrating consistent edits into the 3D model, we observe our approach that converges in significantly fewer iterations compared to GaussianEditor.

\begin{figure}[t]
    \centering
    \includegraphics[width=\linewidth]{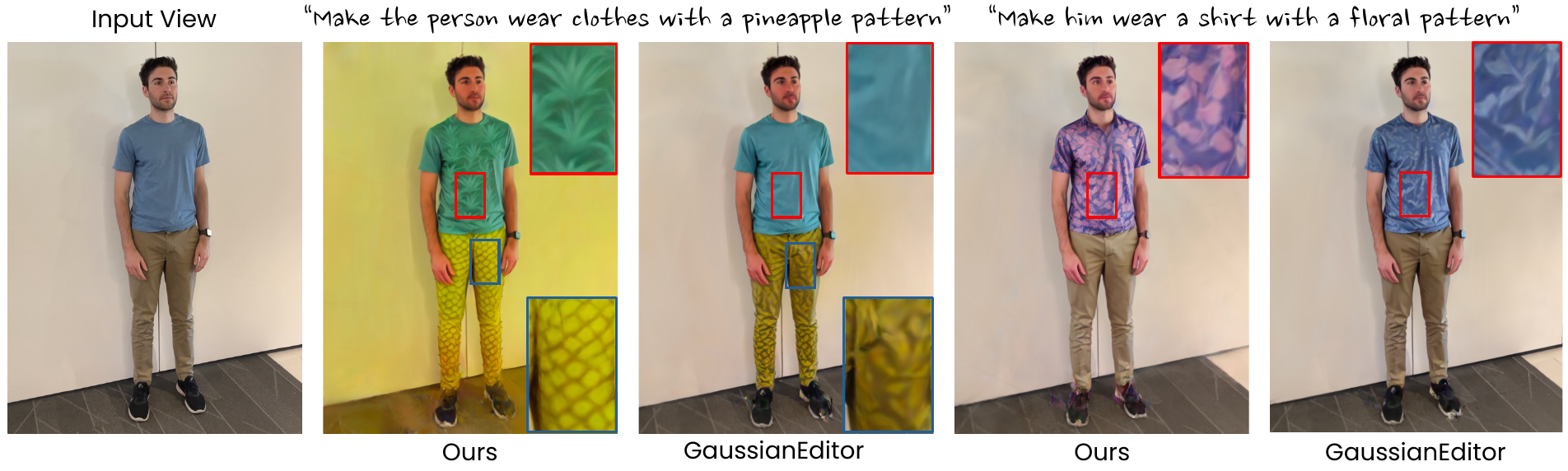}
    \vspace{-0.5em}
    \caption{\textbf{Details Comparison.} \method generates editing results with more detailed textures such as the pineapple or floral pattern.}
    \label{fig:detailed_comparison}
\end{figure}

We provide an additional qualitative comparison with GaussianEditor to assess the edited image details in \Cref{fig:detailed_comparison}. As shown in the figure, our method generates high-fidelity textures that align well with the textual description, which gives the human a pineapple-like outfit.
In contrast, GaussianEditor falls short in delivering this complex pattern with adequate detail. 
This could be attributed to inconsistent editing results of InstructPix2Pix for different views during the process, as also previously noted by IN2N~\cite{haque23instruct-nerf2nerf:}.

\section{Additional Details}
\label{sec:addtional_details}

\subsection{Implementation Details}

\paragraph{Image Editor.} 
We use InstructPix2Pix as the base image editor. 
We follow the same setting in \cite{haque23instruct-nerf2nerf:, chen2023gaussianeditor}, which uses DDIM \cite{song2021denoising} scheduler with 20 steps and the implementation included in the Diffusers \cite{mascaro21diffuser:} library. Different from IN2N \cite{haque23instruct-nerf2nerf:} or GaussianEditor \cite{chen2023gaussianeditor}, which sample noise from $t \in [0.02, 0.98]$, we directly use $t=1$ as we do not require iterative dataset updates. For the refinement step, we add noise with $t=0.3$ to capture more details of 3D editing. We select one keyframe in every five frames. 

\paragraph{Camera Sorting.}
Our approach to multi-view is inspired by video editing. 
While in videos camera trajectories are typically smooth, in our case, when rendering views from the 3D model, we do not assume that a camera trajectory is known in advance.
However, obtaining such a trajectory is useful to ensure that the selected keyframes will cover a variety of viewpoints.  
To this end, prior to multi-view editing, we sort the rendered viewpoints as follows.   
(1) We take the camera with the largest value along the x-axis in world coordinates as the reference camera.
(2) We sort the cameras using the relative angle between their forward vectors and that of the reference camera.
The forward vector is defined as the normalized vector along the z-axis of the camera in world coordinates.

\begin{figure}[t]
    \centering
    \includegraphics[width=\linewidth]{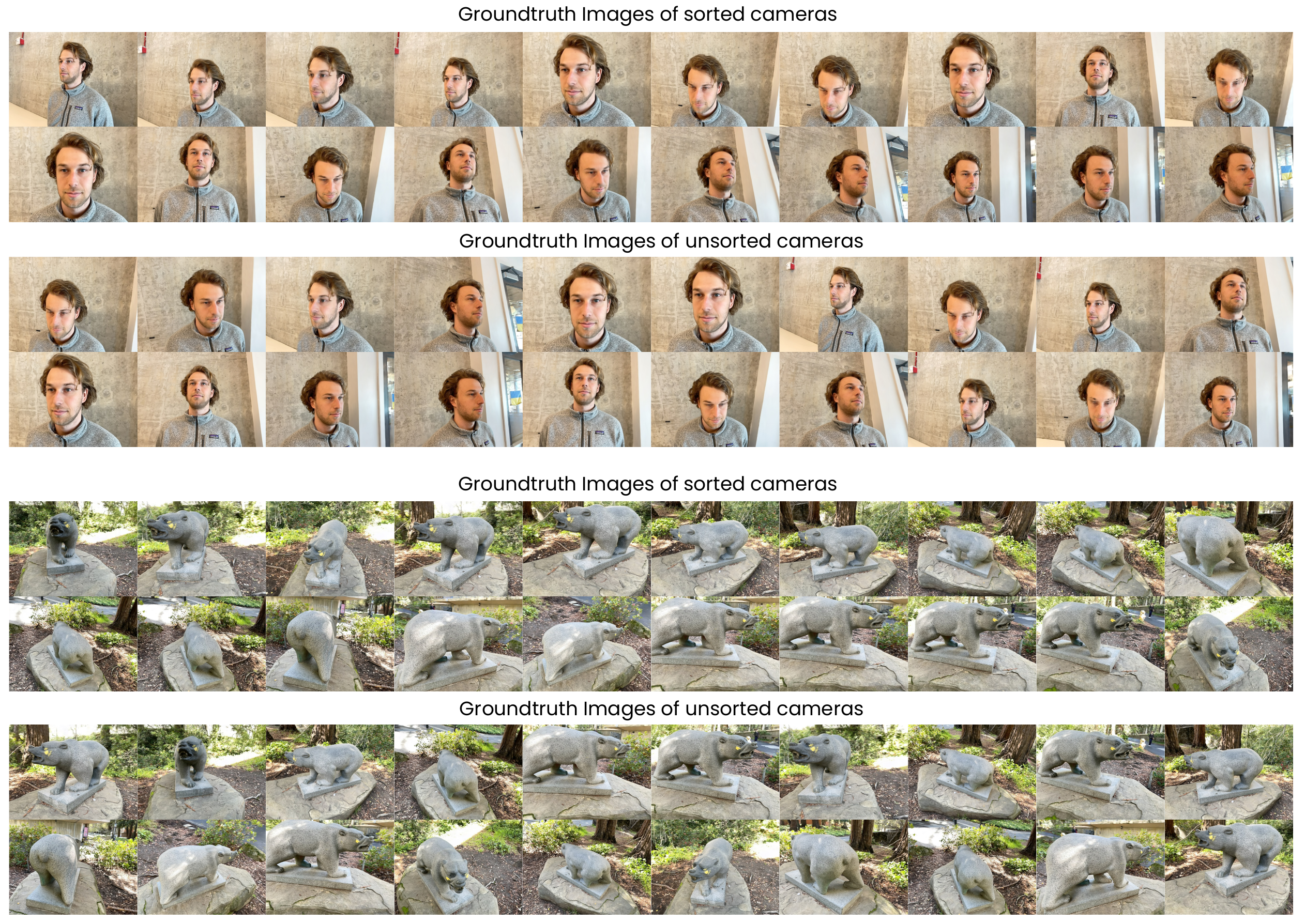}
    \vspace{-0.5em}
    \caption{\textbf{The effect of camera sorting.} The first two rows for each scene are the ground-truth images of sorted cameras and the bottom two rows are ground-truth images of unsorted cameras. Note the smoother transitions from one frame to the next after sorting.}
    \label{fig:camera_sort}
\end{figure}

In \Cref{fig:camera_sort}, we visualize the images corresponding to sorted cameras and unsorted cameras for two different scenes.

\paragraph{Computation Time.}
The average editing time reported (for both DGE and other methods) is evaluated on one Nvidia RTX A6000 GPU with 48G memory. 

\subsection{Evaluation Prompts}
We have created a set of 10 scene-prompt pairs for evaluation purposes, shown in \Cref{table:evaluation_dataset}.
These are used for computing the CLIP and CLIP directional scores. 
We randomly sample 20 views from the training cameras and use the average score as the score for each scene-prompt pair.

\begin{table} [t!]
  \footnotesize
  \newcommand{\xpm}[1]{{\tiny$\pm#1$}}
  \centering
\setlength{\tabcolsep}{2.7pt}
\begin{tabular}{@{}cC{3.5cm}C{3.5cm}C{3.5cm}@{}}
  \toprule
  Scene  & Source Prompt  & Target Prompt  & Edit Instruction  \\
    \midrule
    Face &  ``A man with curly hair in a grey jacket'' & ``A man with curly hair in a grey jacket with a Venetian mask'' & ``Give him a Venetian mask'' \\
    Face &  ``A man with curly hair in a grey jacket'' & ``A man with curly hair in a checkered cloth'' & ``Give him a checkered jacket'' \\
    Face &  ``A man with curly hair in a grey jacket'' & ``A spider man with a mask and curly hair'' & ``Turn him into spider man with a mask'' \\
    Bear &  ``A stone bear in a garden'' & ``A robotic bear in the garden'' & ``Make the bear look like a robot'' \\
    Bear &  ``A stone bear in a garden'' & ``A panda in the garden'' & ``Make the bear look like a panda'' \\
    Bear &  ``A stone bear in a garden'' & ``A bear with rainbow color'' & ``Make the color of the bear look like rainbow''\\
    Person &  ``A man standing next to a wall wearing a blue T-shirt and brown pants'' & ``A man looks like a mosaic sculpture standing next to a wall'' & ``Make the man look like a mosaic sculpture'' \\
    Person &  ``A man standing next to a wall wearing a blue T-shirt and brown pants'' &  ``A man wearing a shirt with a pineapple pattern'' & ``Make the person wear a shirt with a pineapple pattern'' \\
    Person &  ``A man standing next to a wall wearing a blue T-shirt and brown pants'' & ``An Iron Man stands to a wall'' & ``Turn him into Iron Man'' \\
    Person &  ``A man standing next to a wall wearing a blue T-shirt and brown pants'' & ``A robot stands to a wall'' & ``Turn the man into a robot'' \\
  \bottomrule
\end{tabular}
\vspace{0.5em}
\caption{\textbf{Detailed evaluation dataset.}  As shown above, we evaluate different methods on the above 10 different scene-prompt pairs. The source and target prompts are used for computing CLIP and CLIP directional scores. The edit instructions are textual inputs of InstructPix2Pix~\cite{brooks2022instructpix2pix}.}%
\label{table:evaluation_dataset}
\end{table}

\section{Social Impact and Limitations}
\label{sec:social_limitation}
\paragraph{Social Impact.}

In this work, we leverage existing datasets in previous studies \cite{haque23instruct-nerf2nerf:, barron22mip-nerf, mildenhall2019llff}. Our examples may include human subjects for whom consent has been obtained, as outlined in \cite{haque23instruct-nerf2nerf:}. The potential outcomes of our editing results may lead to negative societal impacts if utilized for harmful purposes. We urge users to employ our methodology responsibly, adhering to ethical guidelines, regulations, and relevant laws.

\paragraph{Limitations.} 

The editing performance of our approach is primarily constrained by the capabilities of the underlying image editor, in this case, InstructPix2Pix~\cite{brooks2022instructpix2pix}. Consequently, it inherits limitations from InstructPix2Pix. As such, our method's capability to handle substantial geometric transformations is limited, such as adjusting the position of a person's arm or simulating actions like jumping. We anticipate that incorporating advanced or alternative 2D diffusion priors could pave the way for overcoming these constraints in the future.

\end{document}